\documentclass[12pt]{article}
\usepackage[utf8]{inputenc}
\usepackage[T1]{fontenc}

\usepackage[compress]{cite}
\usepackage{IEEEtrantools,stackengine}
\stackMath

\usepackage{authblk}
\usepackage{fullpage}
\usepackage{amssymb,amsmath}
\usepackage{adjustbox}
\usepackage{graphicx}
\usepackage{longtable}

\usepackage{siunitx}
\usepackage{xcolor}
\usepackage{listings}
\lstdefinestyle{promptstyle}{
  basicstyle=\singlespacing\ttfamily\small,
  breaklines=true,
  breakatwhitespace=false,
  columns=fullflexible,
  keepspaces=true,
  frame=single,
  backgroundcolor=\color{gray!5},
  rulecolor=\color{gray!50}
}

\usepackage{csquotes}

\usepackage{xspace}

\newcommand\eg{e.\,g.\xspace}

\usepackage{booktabs}
\usepackage{longtable}
\usepackage{multirow}

\usepackage[dvipsnames, table]{xcolor}

\definecolor{darkgreen}{rgb}{0.0, 0.5, 0.0}

\definecolor{lightyellow}{HTML}{FFE699}
\definecolor{red_revision}{HTML}{FF0000}

\usepackage{setspace}
\usepackage[unicode=true]{hyperref}
\hypersetup{breaklinks=true,
            pdfborder={0 0 0},
            allcolors=black}
\usepackage[%
  sort&compress
]{cleveref}
  \Crefname{appendix}{Supplement}{Supplements}
  \Crefname{figure}{Fig.}{Fig.}

\usepackage{times}
\usepackage{enumerate}
\usepackage{enumitem}
\usepackage{float}

\usepackage{subfig}
\usepackage{graphicx}

\usepackage{rotating}
\usepackage{tabularx}
\usepackage{dcolumn}
\usepackage{pdflscape}
\usepackage{rotating}

\usepackage{array}

\usepackage{titlesec}

\titleformat{\subsubsection}
  {\normalfont\itshape}{\thesubsubsection}{1em}{}

\titleformat{\subsection}
  {\normalfont\bfseries}{\thesubsection}{1em}{}

\makeatletter
\renewcommand{\fps@figure}{H}         
\renewcommand{\fps@table}{H}         
\makeatother

\allowdisplaybreaks

\usepackage{eurosym}
\usepackage{comment}

\usepackage{ntheorem}
\theoremseparator{:}

\usepackage{makecell}

\begin{document}


\title{\centering\LARGE\singlespacing Automated reproducibility assessments in the social and behavioral sciences using large language models}

\renewcommand\Affilfont{\fontsize{9}{10.8}\selectfont}

\author[1,2,$\dagger$]{Tobias Holtdirk}
\author[1,2,$\dagger$]{Pietro Marcolongo}
\author[1,2]{Anna Steinberg Schulten}
\author[1,2]{Felix Henninger}
\author[3]{Stefan Rose}
\author[1,2]{Sarah Ball}
\author[1,2]{Bolei Ma}
\author[,1,2,3,$\ddagger$]{Frauke Kreuter\thanks{Correspondence: frauke.kreuter@stat.uni-muenchen.de, weinmann@wiso.uni-koeln.de, feuerriegel@lmu.de}}
\author[*,4,$\ddagger$]{Markus Weinmann}
\author[*,1,2,$\ddagger$]{Stefan Feuerriegel}


\affil[1]{LMU Munich, Munich, Germany}
\affil[2]{Munich Center for Machine Learning, Munich, Germany}
\affil[3]{University of Maryland, College Park, USA}
\affil[4]{University of Cologne, Cologne, Germany}
\affil[$\dagger$]{Joint first authorship}
\affil[$\ddagger$]{Joint last authorship}

\date{}

\maketitle

\newpage

\begin{abstract}\normalfont
\noindent
Reproducibility in the social and behavioral sciences is typically evaluated by independent researchers who reanalyze the original data to assess whether the published findings can be recovered. However, such approaches are resource-intensive and difficult to scale. Here, we show that large language models (LLMs) can automate reproducibility assessments. Using $N = 180$ published studies with predefined claims from the behavioral and social sciences, we compare LLM-generated analyses with the original findings. For 11 studies, the LLM pipeline could not produce a viable effect size estimate. For the remaining studies, the LLM reached the same qualitative conclusion as the original study in 80\% of cases, and recovered the original effect sizes (using a $\pm$0.05 tolerance in Cohen's $d$) in 24\% of studies. In a subset with human reanalyses, the LLM reached the same qualitative conclusion as the original study in 95\% of studies, similar to human reanalysts (83\%), and the LLM recovered the original effect sizes using a $\pm$0.05 tolerance in 40\% of studies, again broadly similar to human reanalysts (28\%). Given the current capabilities and limitations of LLMs, the findings show that LLMs can support systematic audits of empirical results rather than substitute expert judgment. As such, LLMs can serve as a scalable screening tool to improve the rigor and reproducibility in empirical research.
\end{abstract}

\flushbottom
\maketitle
\thispagestyle{empty}


\sloppy
\raggedbottom


\newpage
\section*{Main}
\label{sec:introduction}


Scientific progress depends on reproducible findings. Yet, across the social and behavioral sciences, reanalyses of original data have shown that published results are not always recovered \cite{Miske.2026,Aczel.2026,Brodeur.2026,Hardwicke.2021, Brodeur.2024}. These patterns make it difficult to know which published findings can be trusted, motivating large-scale efforts to reanalyze published studies and check whether their central findings can be reproduced from the original data. Ideally, journals would reanalyze submissions before publication to catch irreproducible results early, and some have started to do so \cite{Fisar.2024,Miske.2026,Hardwicke.2021,Nosek.2026}.


However, reanalyzing published studies is difficult to scale because it requires a time-intensive reconstruction of the original empirical workflow \cite{Aczel.2026, Miske.2026, Brodeur.2026, Hardwicke.2021, Brodeur.2024}. Unlike replication efforts, which test whether a finding holds in newly collected data \cite{Tyner.2026, Parsons.2022, OSC.2015}, reproducibility concerns whether a reported result can be recovered from the original data and study materials \cite{Peng.2011,Parsons.2022}.\footnote{We use ``reproducibility'' throughout in this broad sense, spanning both close reimplementation and claim-level robustness, and our design varies how much of the original methodology the model receives. The two ends place different demands on the analyst and imply different benchmarks: the original reported result for close reproduction, and the distribution of defensible reanalyses for robustness.} For this, analysts must inspect the study materials, identify the relevant files and variables, prepare the data, translate the paper's description into executable code, run the analysis, and compare the resulting statistic with the original finding. The scale of this effort is demonstrated by the recent Multi100 collaboration, wherein 507 analysts reanalyzed 100 published studies in economics, political science, and psychology over several years \cite{Aczel.2026}. Overall, reproducibility assessments are resource-intensive and study-specific, which makes systematic auditing of the empirical literature costly and difficult to scale.


One way to make reproducibility assessments scalable is to automate the reanalysis itself. Recent advances in large language models (LLMs) have produced systems with strong coding capabilities \cite{Sun.2026} and growing use in scientific programming, data analysis, and research assistance \cite{Boiko.2023, Qian.2024, Lu.2026, Schmidgall.2025, Seo.2026, Alizadeh.2026, Kohler.2026, Miao.2025, Song.2025, Shao.2026, Zhang.2025, Gottweis.2026, Ghareeb.2026, Yamada.2025}. Here, we propose using LLMs as automated analysts for reproducibility assessment. In principle, such systems can process study materials and generate executable analysis code, which could reduce the extensive manual effort currently required for data and reproducibility checks \cite{Fisar.2024} and make such checks feasible across a broader range of publication outlets as part of routine quality control. However, LLM-based reanalysis is challenging, since the generated code may contain errors, it is unclear how accurately LLMs can interpret the analytical decisions implied by the study materials, and outputs may reflect unsupported assumptions about the data \cite{Song.2026}. Hence, whether LLMs can reliably carry out the computational reproducibility assessments is unknown.

\begin{figure}
\centering
\includegraphics[width=1\linewidth]{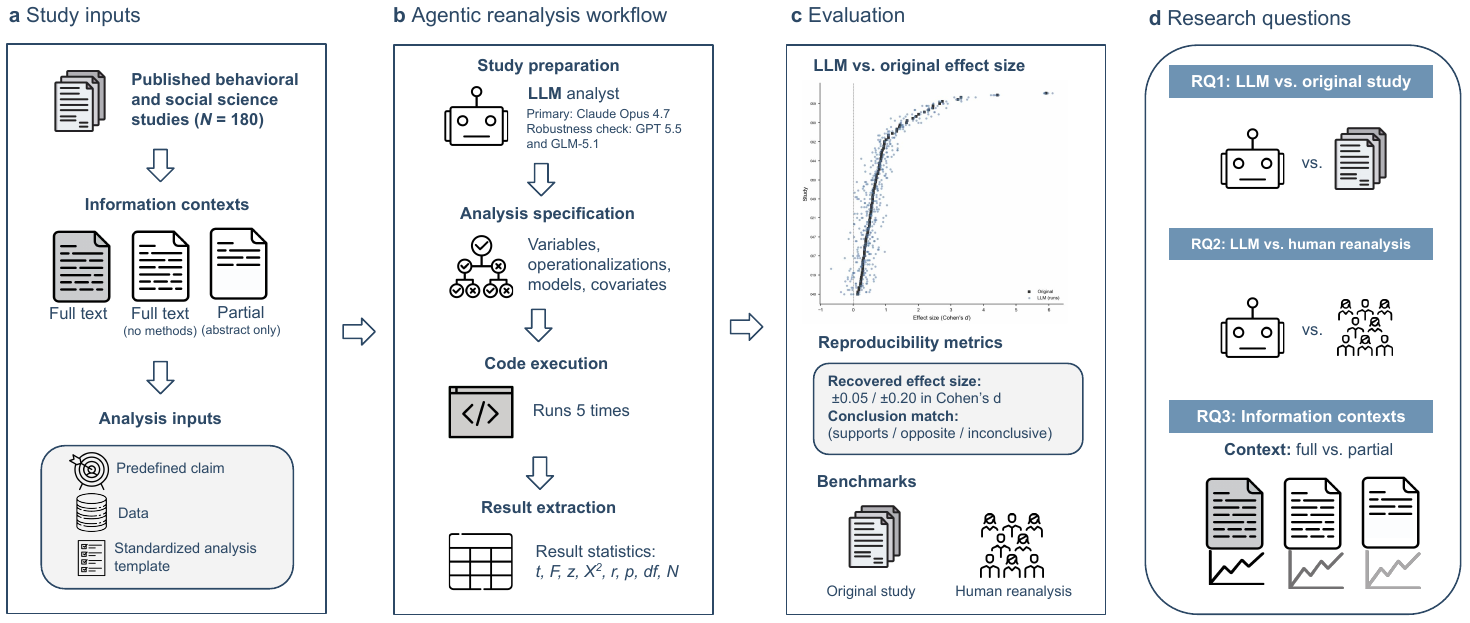}
\caption{\textbf{Automated reproducibility assessment using LLMs.} \textbf{a,}~A corpus of published studies ($N = 180$) with predefined claims, datasets, and standardized analysis templates is provided across varying information contexts (full text, full text without methods, or abstract only). \textbf{b,}~The agentic LLM pipeline processes study materials, specifies analytical choices (including variables, operationalizations, models, and covariates), and extracts the resulting statistics (over five independent runs). \textbf{c,}~Extracted effect sizes and conclusions are evaluated against the reported results (from the original study) using several metrics to assess the reproducibility, namely, whether the effect size is recovered (within a $\pm0.05$ and $\pm0.20$ Cohen’s $d$ tolerance) and whether the qualitative conclusions match (i.e., support / opposite / inconclusive). \textbf{d,}~We analyze three main research questions: (1)~the overall reproducibility of findings, (2)~the reproducibility compared to human analysts, and (3)~the analytical variability due to different information contexts.
}
\label{fig:overview}
\end{figure}

Here, we examine whether LLMs can enable automated reproducibility assessments in the social and behavioral sciences (see \Cref{fig:overview}). We develop an agentic LLM pipeline in which the model receives the original dataset, a focal statistical claim, and experimentally varied contextual information from the article, and is tasked to independently write and execute statistical code to reproduce the claims. Across a preregistered corpus of published empirical studies from psychology, political science, and economics, we address the following research questions. (1)~To what extent can LLM-generated analyses reproduce the statistical results and substantive conclusions of published studies? To this end, we compare standardized effect sizes (Cohen's $d$) with the original published findings and further assess whether the LLM recovers the same substantive conclusion. (2)~How do the LLM-generated reproducibility analyses compare against human reanalyses? For a subset of papers, we additionally compare the LLM-generated effect sizes against those from a large-scale human reanalysis effort \cite{Aczel.2026}. This comparison is descriptive rather than a direct performance benchmark; the human reanalysts may have pursued their own defensible specifications rather than aiming to reproduce the originally reported result, whereas the LLM pipeline was instructed to test the focal claim using the paper and available data. Therefore, higher agreement between the LLM and the original findings may reflect a narrower or more paper-anchored analytical path rather than greater accuracy relative to human reanalysis. (3)~How does reproducibility vary with the amount of methodological information provided to the model, that is, with access to the full method description versus a more conceptual reanalysis where only the abstract of the paper is provided? For our main analysis, we use Claude Opus~4.7 as a state-of-the-art LLM for code generation, and demonstrate the generalizability using two additional frontier models, namely, GPT-5.5 (from OpenAI) and GLM-5.1 (an open-weight model from Zhipu).

\newpage

\section*{Results}

\subsection*{Study overview}


We analyzed $N=180$ studies from the Systematizing Confidence in Open Research and Evidence (SCORE) project \cite{Alipourfard.2021}. SCORE is a large-scale research initiative in the social and behavioral sciences in which published empirical claims were identified and independently evaluated. We used studies for which focal claims had already been extracted and for which the original study data had been retrieved and made available, or for which data from a replication effort were available. We evaluated the automated reanalysis using the following metrics \cite{Aczel.2026}: (1)~the difference in effect size compared to the original finding, defined as $\Delta d = \bar{d}_{\mathrm{LLM}} - d_{\mathrm{original}}$, where $d_{\mathrm{original}}$ is the original effect size and where $\bar{d}_{\mathrm{LLM}}$ is the mean effect size of the LLM reanalysis over five runs, (2)~whether the result fell within a tolerance region of $\pm$0.05 (strict) or $\pm$0.20 (broad), and (3)~whether the substantive conclusion matched the original claim (classified as support / opposite / inconclusive; based on the majority vote across runs). 

We evaluated each study using the agentic LLM pipeline across five independent runs to address variability in LLM outputs. In each run, the model received the focal claim, the available study materials (i.e., the entire paper), and the corresponding data as input, and was asked to write and execute statistical code to estimate the effect directly tied to the claim. The model was allowed to make reasonable analytical choices where the claim or study materials were underspecified, but we included safeguards to prevent direct copying from the paper; for example, the model was instructed to compute the statistic from code executed on the data, not to reuse reported test statistics, and to return a structured output containing the estimated statistic, sample size, degrees of freedom, $p$-value, effect size information, and qualitative conclusion. The LLM pipeline was allowed to install additional software packages, if needed. Claude Opus 4.7 served as the primary LLM, while we later also test the generalizability using GPT-5.5 and GLM-5.1.

Overall, 11 studies were excluded because none of the five runs yielded a valid Cohen's $d$, yielding $N=169$ papers for our analysis.

\subsection*{LLMs can automate reproducibility assessment (RQ1)}

\begin{figure}
\centering
\includegraphics[width=0.95\linewidth]{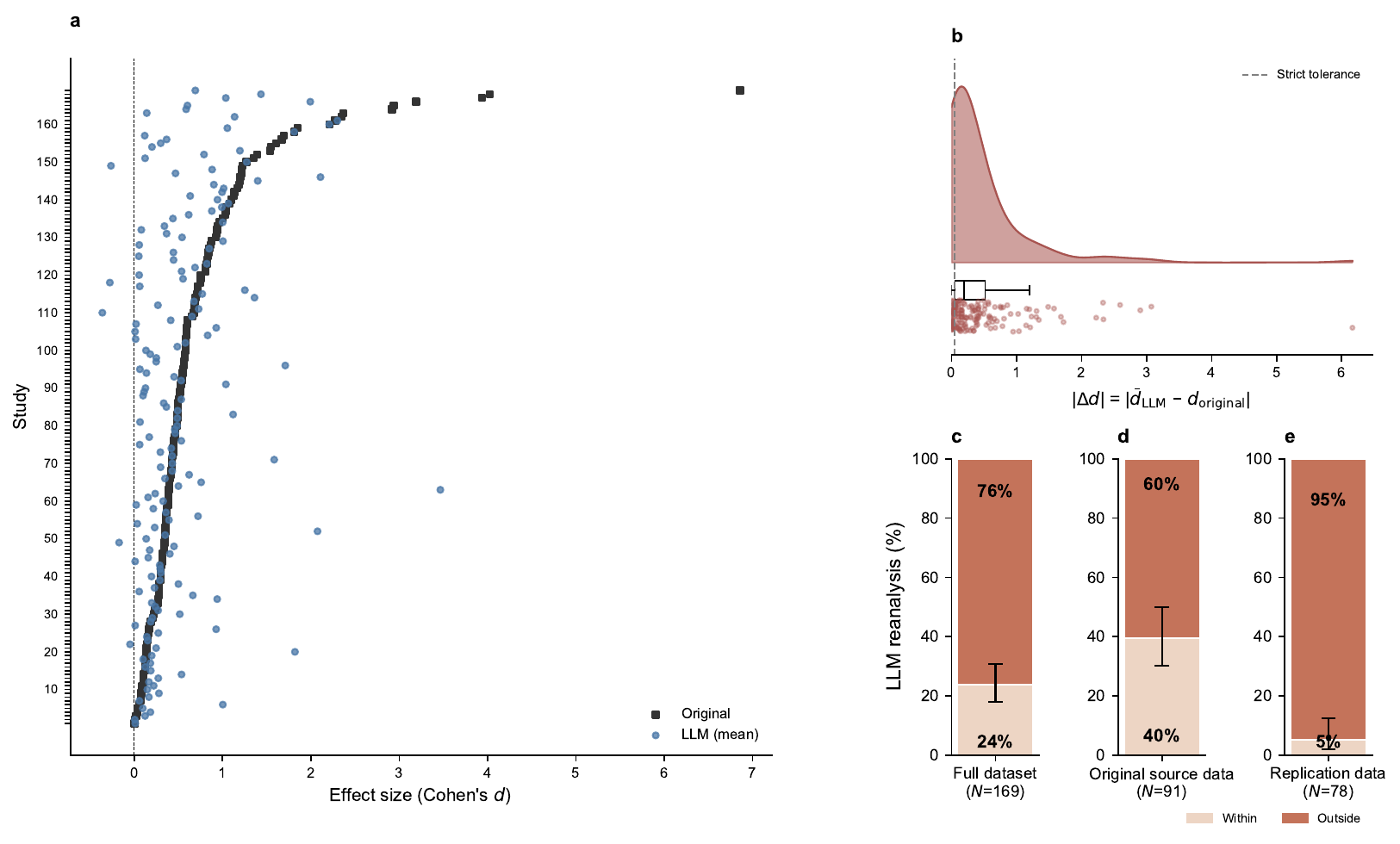}
\caption{\textbf{Automated reproducibility assessment (using Claude Opus 4.7).} 
\textbf{a,}~Effect size of the original analysis (gray squares; all represented as positive values) and the effect sizes of the reanalyses (blue dots) for each study. Shown are the $N = 169$ studies for which a valid Cohen's $d$ was produced by the LLM (while 11 studies were excluded for that reason). 
\textbf{b,}~Distribution of $|\Delta d|$, computed as $|\Delta d| = |\bar{d}_{\mathrm{LLM}} - d_{\mathrm{original}}|$.
The distribution is visualized using a density plot, a boxplot, and scatter points. In the boxplot, the line indicates the median, the box denotes the interquartile range (IQR), and the whiskers extend to 1.5$\times$ of the IQR; points beyond the whiskers indicate outliers.
\textbf{c,}~Proportion of studies for which the LLM-generated effect size falls within or outside the tolerance region around the original result, across the full sample.
\textbf{d,}~Proportion of studies falling within or outside the tolerance region, for the subset of studies that make the original source data available.
\textbf{e,}~Proportion of studies for which the LLM-generated effect size falls within or outside the tolerance region, for the subset of studies where the original source data were unavailable but where replication data for reanalyses are provided. 
Whiskers indicate Wilson 95\% confidence intervals.
}
\label{fig:rq1_main_result}
\end{figure}

The distribution of estimated effect sizes based on the LLM reanalysis is shown in \Cref{fig:rq1_main_result}a. The difference in effect size compared to the original finding is shown in \Cref{fig:rq1_main_result}b. The distribution is highly right-skewed, with most studies showing relatively small deviations from the original effect size but a small number of studies exhibiting large discrepancies. We next compare the mean effect size from the LLM reanalysis to the original published effect size (\Cref{fig:rq1_main_result}c). The LLM reanalysis fell within the strict tolerance of $\pm 0.05$ for 24\% of the studies (22\% when treating studies without a valid LLM effect size estimate as non-recovered). Using the broader tolerance of $\pm 0.20$, the LLM reanalysis fell within tolerance for 50\% of the studies (47\% when treating studies without a valid LLM effect size estimate as non-recovered; Supplementary~Fig.~\ref{suppfig:rq1_main_result_broad}). Together, these results show that the LLM pipeline was able to recover the original findings in a substantial share of cases.


We further examined whether reproducibility differed between studies using the original source data (named `reproduction data' in SCORE) versus studies where the original source data were unavailable but where replication data were made available for reanalysis (\Cref{fig:rq1_main_result}d,e). Reproducibility was higher for studies using source data than for studies using replication data, consistent with prior findings that replication studies collecting new data to test whether an effect holds have repeatedly recovered weaker effects than originally reported \cite{OSC.2015,Tyner.2026}.

\newpage

\subsection*{Comparison of LLM-powered vs. human reanalyses (RQ2)}

Next, we compare the mean effect size obtained from the LLM reanalysis to the effect size from human reanalyses. We thus focus on the subset of $N = 84$ papers for which such human benchmarks have been generated over the course of multiple years as part of a large-scale human reanalysis effort \cite{Aczel.2026}. Out of the $N = 84$ studies, the LLM pipeline did not produce a valid Cohen's $d$ for two studies, leaving $N = 82$ studies for the comparison. 

\begin{figure}
\centering
\includegraphics[width=0.6\linewidth]{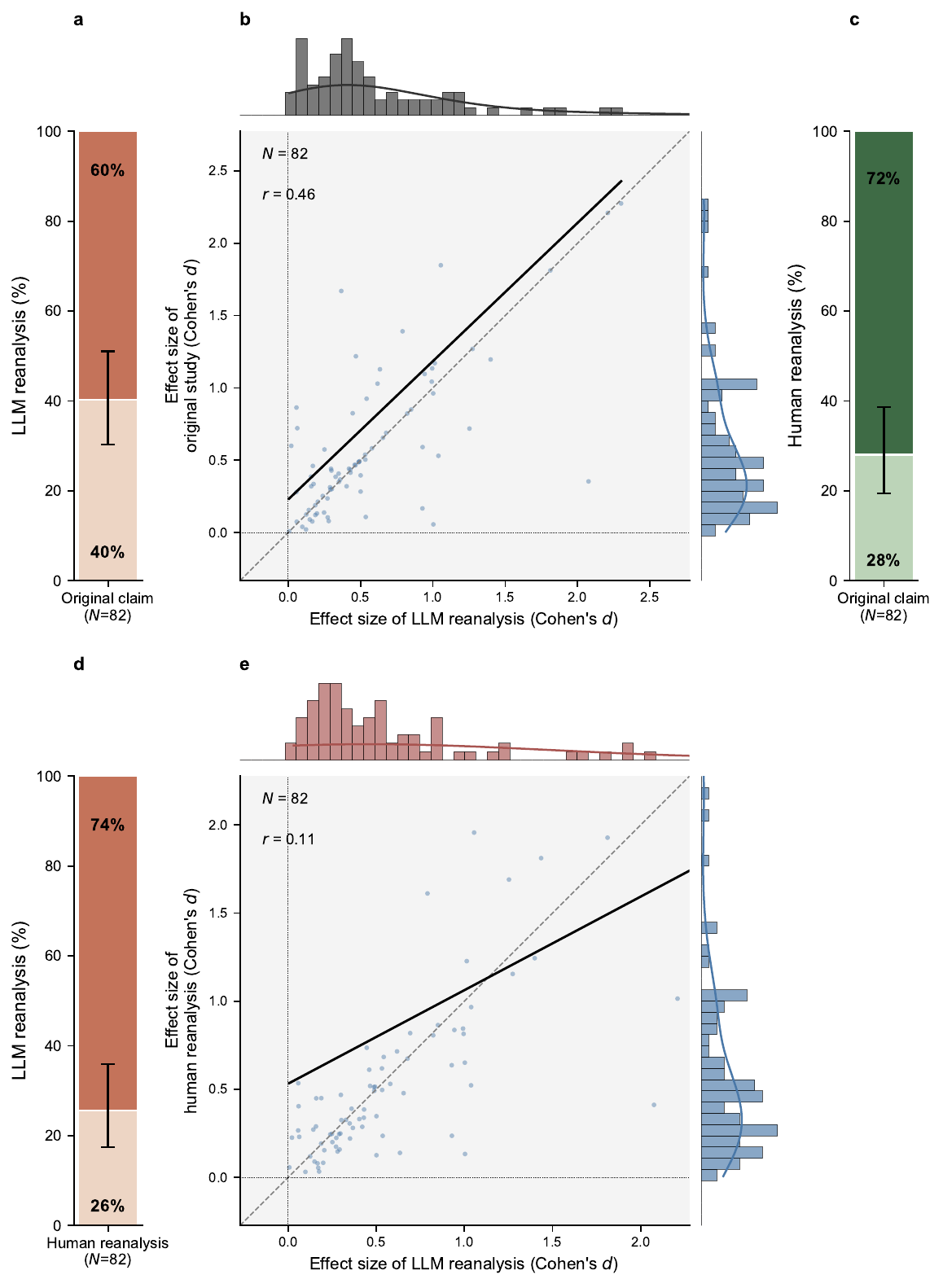}
\caption{\textbf{LLM reproducibility compared with human reanalyses.} 
We compare Cohen's $d$ from the automated LLM reanalysis with the original published effect sizes (\textbf{a,b}) and with effect sizes from human reanalyses (\textbf{d,e}). Here, we focus on the subset of studies for which human reanalysis benchmarks were available from a large-scale reanalysis effort \cite{Aczel.2026}. To compare the LLM performance, we also report the human reanalysis results for recovering the original effect sizes (\textbf{c}). Results ($N = 82$ studies) are based on Claude Opus~4.7. 
\textbf{a,}~Proportion of studies for which the LLM-generated effect size falls within or outside the tolerance region around the original published effect size.
\textbf{b,}~Scatterplot comparing the Cohen's $d$ from the LLM reanalysis against the Cohen's $d$ from the original study. The thin diagonal line represents the ideal case in which the reanalysis effect sizes are equal to the original effect size. The thick line is the estimated trend line ($r$ is the Pearson correlation coefficient). Density plots on the axes show the respective distribution of the effect sizes. 
\textbf{c,}~Proportion of studies for which the human reanalysis effect size falls within or outside the tolerance region around the original published effect size.
\textbf{d,}~Proportion of studies for which the LLM-generated effect size falls within or outside the tolerance region around the human reanalysis.
\textbf{e,}~Scatterplot comparing Cohen's $d$ from the LLM reanalysis against the Cohen's $d$ from the human reanalyses (averaged over all reanalyses per each study).
Whiskers indicate Wilson 95\% confidence intervals. 
}
\label{fig:rq2_effect_size_scatter}
\end{figure}

For this subset of studies, the LLM reanalysis fell within the strict tolerance of $\pm 0.05$ in 40\% of studies (\Cref{fig:rq2_effect_size_scatter}a). This is higher than the corresponding rate in the full sample and is explained by the composition of the subset, in which the original source data were available for most studies. We then compared the LLM reanalysis with the human reanalysis benchmark. The human reanalyses fell within the same strict tolerance of the original effect size in 28\% of studies (\Cref{fig:rq2_effect_size_scatter}c). Using the broader $\pm0.20$ tolerance, the LLM reanalysis fell within tolerance in 65\% of studies, compared with 66\% for the human reanalyses. The substantive conclusion from the LLM reanalysis matched the original claim in 95\% of studies, while the substantive conclusion from the human reanalysis matched it in 83\%.

To understand whether the LLM tends to recover the effect sizes of the original paper vs. the human reanalyses, we plotted the LLM-derived effect sizes against the original published effect sizes (Fig.~\ref{fig:rq2_effect_size_scatter}b) and against the human reanalysis effect sizes (Fig.~\ref{fig:rq2_effect_size_scatter}e). The correlation between the LLM reanalysis and the original study effect sizes was moderate ($r = 0.46$; $p < 0.001$; Fig.~\ref{fig:rq2_effect_size_scatter}b), indicating that the LLM captured part of the overall variation in the original effects despite substantial study-level variation. By contrast, the correlation between the LLM-derived effect sizes and the human reanalysis effect sizes was weak ($r = 0.11$; $p = 0.31 $; Fig.~\ref{fig:rq2_effect_size_scatter}e), suggesting that the LLM estimates aligned more closely with the original published effects than with the human reanalysis estimates.

We next assessed whether the LLM-derived effect sizes tracked variation in the original published effect sizes and in the human reanalysis effect sizes. To do so, we estimated study-level linear regressions. Regressing the original published effect size $d_{\mathrm{original}}$, as dependent variable, on the mean LLM-derived effect size $\bar{d}_{\mathrm{LLM}}$, as independent variable,  yielded a slope of $\beta = 0.96$ (SE = 0.21, 95\% CI [0.55, 1.37], $p<0.001$) and explained 21\% of the variance ($R^2 = 0.21$). For comparison, regressing the human reanalysis effect size $\bar{d}_{\mathrm{human}}$, as dependent variable, on the mean LLM-derived effect size $\bar{d}_{\mathrm{LLM}}$, as independent variable, yielded a slope of $\beta = 0.53$ (SE = 0.52, 95\% CI [$-$0.50, 1.56], $p = 0.31$) and explained 1\% of the variance ($R^2 = 0.01$). This indicates that the LLM-derived effect sizes tracked the original published effect sizes more closely than the human reanalysis effect sizes.

\newpage
\thispagestyle{empty}

\clearpage


Multi-analyst and multiverse studies have shown that different defensible analytical choices can yield substantial variation in effect size estimates, even when analysts start from the same data and hypothesis \cite{Silberzahn.2018,BotvinikNezer.2020,Steegen.2016,Breznau.2022,Aczel.2026}. To assess whether similar analytical variability arises in automated reanalysis across different runs of the LLM pipeline, we compared the minimum-to-maximum range of human reanalysis estimates with the corresponding minimum-to-maximum range of LLM-generated estimates for the same studies (Supplementary~Fig.~\ref{suppfig:effect_size_ranges}). The range of LLM-generated estimates generally followed a pattern similar to the range of human reanalysis estimates, suggesting that LLM reanalyses may capture part of the same study-level analytical variability observed among human analysts.

\subsection*{Sensitivity to different information contexts (RQ3)}

\begin{figure}
\centering
\includegraphics[width=0.95\linewidth]{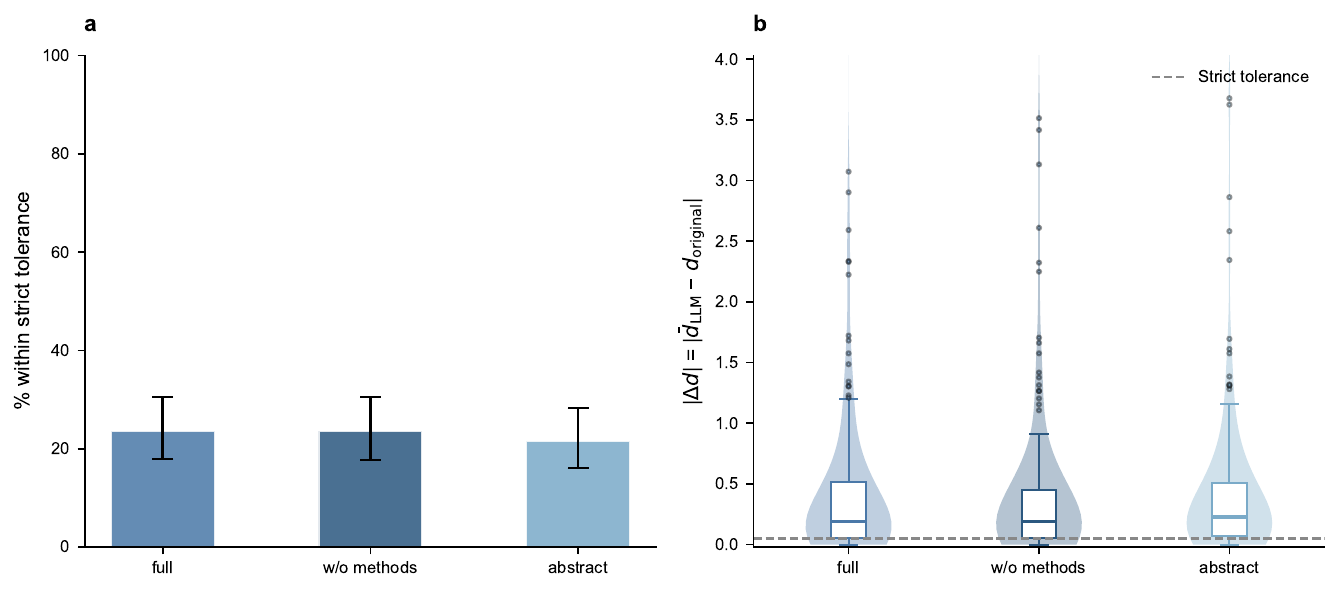}
\caption{\textbf{Sensitivity to different information contexts.}
We compared three input variants (using Claude Opus~4.7): the full paper, the full paper with the methods section removed, and the abstract only.
\textbf{a,}~Proportion of studies where the mean LLM effect size falls within $\pm$0.05 Cohen's $d$ of the original, by analysis variant. Whiskers show Wilson 95\% confidence intervals. \textbf{b,}~Distribution of absolute effect size deviations (i.e., $|\Delta d| = |\bar{d}_{\text{LLM}} - d_{\text{original}}|$) by variant. The dashed line shows the $\pm$0.05 tolerance threshold. In the boxplot, the line indicates the median, the box denotes the interquartile range (IQR), and the whiskers extend to 1.5$\times$ of the IQR; points beyond the whiskers indicate outliers.
}
\label{fig:rq3_info_condition}
\end{figure}

To examine whether reproducibility varied with the amount of article context provided to the LLM, we compared three input variants: the full paper, the full paper with the methods section removed, and the abstract only. Reproducibility rates were similar across these conditions. The LLM-generated effect size fell within the strict $\pm0.05$ tolerance in 24\% of studies in the full paper condition, 24\% in the no methods condition, and 22\% in the abstract-only condition (\Cref{fig:rq3_info_condition}a). Using the broader $\pm0.20$ tolerance, the corresponding rates were 50\%, 52\%, and 46\% for the full paper, no methods, and abstract-only conditions, respectively. The deviations in the absolute effect size (i.e., $|\Delta d|$) were also comparable across conditions (\Cref{fig:rq3_info_condition}b).

We formally tested differences across information conditions using Cochran's $Q$ tests with Bonferroni-corrected post-hoc McNemar tests for the binary within-tolerance outcome, and Friedman tests with Bonferroni-corrected post-hoc Wilcoxon signed-rank tests for absolute effect size deviations. These tests showed no evidence of statistical differences across conditions (Cochran's $Q = 0.80$, $p = 0.67$, and Friedman $\chi^2 = 2.87$, $p = 0.24$, with no significant pairwise contrasts after correction).

\subsection*{Robustness across different LLMs}

To demonstrate the generalizability of our findings across different LLMs, we next repeated our analysis with different LLMs. In addition to the primary analyses with Claude Opus~4.7, we performed the automated reproducibility assessment using GPT-5.5 and GLM-5.1. Overall, the results were largely similar. The mean LLM effect size fell within the strict $\pm 0.05$ Cohen's $d$ tolerance region of the original published effect size in 24\% of studies for Claude Opus~4.7, 34\% for GPT-5.5, and 31\% for GLM-5.1 (Supplementary~Fig.~\ref{suppfig:rq1_main_result_gpt} and \ref{suppfig:rq1_main_result_glm}). Using the broader $\pm 0.20$ tolerance region, the corresponding reproducibility rates were 50\%, 58\%, and 55\%, respectively (Supplementary~Fig.~\ref{suppfig:rq1_main_result_broad}, \ref{suppfig:rq1_main_result_gpt_broad}, and \ref{suppfig:rq1_main_result_glm_broad}). The qualitative conclusion matched the original conclusion in 80\% of studies for Claude Opus~4.7, 81\% for GPT-5.5, and 83\% for GLM-5.1. Together, these results indicate that the main findings are not specific to a single LLM, but generalize across different frontier models, including an open-weight model.

\subsection*{Heterogeneity analysis}

To assess the sensitivity to the prompt framing, we compared the neutral baseline prompt with two alternative prompt framings, namely, a confirmatory framing that encouraged the model to provide supportive evidence of the claim, and a critical framing that encouraged a skeptical test. Differences across confirmatory and critical prompt framings were small (i.e., the strict within-tolerance rate differed by only 4 percentage points between the confirmatory and critical framings; Supplementary~Fig.~\ref{suppfig:perspective}).

Finally, we analyzed the heterogeneity across different study characteristics. First, we compared the reproducibility rates across experimental and observational studies to assess whether reproducibility differed by empirical design (Supplementary~Fig.~\ref{suppfig:by_study_characteristics}). Second, we stratified studies by public code availability to test whether the results depended on the transparency of the original analysis materials (Supplementary~Fig.~\ref{suppfig:code_availability}). Third, we analyzed reproducibility separately for psychology, economics, and political science to account for domain-specific differences in data structures, statistical models, and reporting conventions (Supplementary~Fig.~\ref{suppfig:by_discipline}). Fourth, we further stratified studies by the magnitude of the original Cohen's $d$ to assess whether larger published effects showed lower reproducibility rates (Supplementary~Fig.~\ref{suppfig:es_magnitude}). The latter analysis addresses the possibility that larger original effects may partly reflect opportunistic analytical choices, selective reporting, or other forms of study-specific flexibility.

\section*{Discussion}


LLMs can reproduce a substantial share of published findings in the social and behavioral sciences. Across the 169 of the 180 studies where the LLM pipeline successfully recovered effect sizes, the estimates were within $\pm$0.05 tolerance of the original Cohen's $d$ in 24\% of studies and reached the same qualitative conclusion as the original study in 80\% of cases. In the subset with human reanalysis benchmarks, the LLM pipeline performed comparably to human reanalysts, where it recovered the original effect sizes in 40\% of studies, broadly similar to 28\% for humans, and produced the original qualitative conclusion in 95\% of cases (83\% for humans). These findings were robust across different LLMs, including open-weight models. At the same time, some runs did not yield usable results, highlighting that LLM-based end-to-end statistical coding and analysis remains an emerging capability and is not yet fully reliable. Nevertheless, these results suggest that LLMs can provide a scalable first-pass tool for reproducibility assessment.


LLM analysis often reached the same substantive answer as the original study, even when it did not recover the exact effect size. The differences between the LLM-generated estimates and the original findings could be due to two main sources. First, such differences may reflect errors or limitations in the LLM-generated analysis. Second, they may also reflect conceptual ambiguity in the original claim, theory, or methods \cite{Auspurg.2021}. Research hypotheses are often short verbal statements that do not uniquely specify the sample, variables, model, or controls, and some analytical strategies are not fully specified, which can create legitimate variation in reanalysis decisions \cite{Scheel.2022}. Consistent with this, the observed distribution of effect sizes appears to be broadly similar between human and automated reanalyses (Supplementary~Fig.~\ref{suppfig:effect_size_ranges}). Distinguishing such variation from modeling errors is itself a known challenge, even for human reanalysis. 

We do not interpret this comparison as evidence that either LLMs or humans are generally superior. Human reanalyses in \cite{Aczel.2026} allowed analysts to pursue their own defensible specifications, while our automated pipeline may have encouraged prioritizing a narrower analytical path that may have been closer to the originally reported result. The latter is reflected in a higher agreement of the LLM estimates with the original effect sizes ($r = 0.46$) than with the human estimates ($r = 0.11$), which should be interpreted cautiously, especially given the limited benchmark subset ($N = 82$). Further, because both human and automated reanalyses involve judgment over operationalizations, samples, models, and covariates, differences in recovered effect sizes may reflect alternative but reasonable analytical choices rather than clear errors by one type of analyst.


Our results also show that LLM-based reproducibility did not meaningfully differ when more or less detailed methodological information was provided (see RQ3). In principle, brief claims leave many analytical choices open (e.g., how theory-laden concepts are operationalized or how hypotheses are tested \cite{Oberauer.2019}). This pattern is consistent with the idea that empirical results are dependent on the particular choices researchers make among many similarly acceptable data processing and modeling choices \cite{Coretta.2023, Gelman.2013, BotvinikNezer.2020, Patel.2015}. One might therefore expect that providing more detailed methodological information would constrain the model’s choices and move its analysis closer to the original. Against this background, the limited variation across information conditions suggests that the model may be guided primarily by the focal claim and the structure of the available dataset. Once the model identifies one plausible mapping between the claim and the data, additional methodological detail appears to have only limited influence on the selected operationalization, model specification, and resulting conclusion. Still, caution is needed when interpreting this finding. Limited variation across information conditions does not imply that analytical choices are unimportant, nor that the model has exhaustively searched the space of defensible analyses (e.g., explicit multiverse analysis \cite{Steegen.2016, Silberzahn.2018, Wagenmakers.2022, Aczel.2026} could still reveal a broad range of possible estimates and conclusions). 


Our study has several strengths. First, the results were robust across three frontier LLMs, namely, Claude Opus 4.7, GPT-5.5, and GLM-5.1. A notable strength is that one of these models was an open-weight LLM, showing that automated reanalysis is not limited to closed proprietary systems. Second, we analyze 180 studies, which exceeds the size of the individual corpora used in recent human reanalysis efforts in this area \cite{Aczel.2026,OSC.2015,Brodeur.2026}. Third, our pipeline is multimodal and can process not only text and tabular data, but also figures and other image-based information extracted from the original papers. 


Practically, automated reanalysis could reduce the burden of data-quality and reproducibility checks that journals increasingly require, and could extend such checks to settings where full manual review is infeasible. Rather than replacing expert judgment, LLM-based reanalysis could support editors, reviewers, and meta-scientists by providing a scalable first-pass check of whether reported findings can be recovered from the available data and study materials. Such workflows could be integrated into submission, post-acceptance, or post-publication processes for quality control and help make reproducibility assessment more routine, especially as some journals have already adopted such initiatives \cite{Fisar.2024,Miske.2026,Hardwicke.2021,Nosek.2026}. More broadly, this approach follows a growing trend toward automated tools to support scientific workflows \cite{Nuijten.2020,Bertran.2026,Nosek.2026,Brodeur.2025, Shao.2026,Siegel.2024,Starace.2025,Wrightson.2025} and positions LLM-based automated reanalysis within the broader movement toward transparency and openness in science \cite{Nosek.2018}, adding a scalable instrument to a toolkit that has so far relied largely on manual, labor-intensive approaches. However, automated checks should not be treated as end-to-end verification, given that reproducibility assessments often depend on many defensible analytical choices, contextual interpretation, and the quality of the available materials. Caution is also needed for two reasons. First, such automated checks may create Goodhart-style incentives, where authors optimize submissions to pass automated screens rather than to improve the underlying analysis or ensure genuinely robust results. Second, LLM-based analysis tools could facilitate gaming behavior by making it easier for researchers to opportunistically screen many analytical specifications \cite{Brodeur.2020,Simmons.2011}, potentially biasing results toward desired findings.


Several limitations apply. First, the performance may depend on the modeling choices, such as the model and prompt. Our sensitivity analyses suggest that the main results are largely stable across these choices. Second, a key concern is training data contamination. Because the studies in our corpus were published before the training cutoff of the models, LLMs may have encountered some of the studies during pretraining, which could inflate reproducibility rates. A related concern applies to the human reanalysis benchmark. To assess this risk, we conducted a memorization test \cite{Nori.2023, Sainz.2023} and further manually inspected the reasoning traces for signs of direct copying or other forms of apparent cheating, but did not find evidence of such. We also compared results based on whether the original analysis code was publicly available, but found similar results. Nevertheless, the absence of direct memorization does not rule out subtler effects of training data exposure (e.g., even if a model cannot recall a reported effect size, prior exposure to a study could still improve the ability to interpret the prompt or write suitable analysis code). Third, our corpus is limited in size and to studies from psychology, economics, and political science. This reflects the difficulty and cost of assembling such datasets; even though our dataset is larger than others, this is a constraint that affects reproducibility audits broadly \cite{Krahmer.2026}. Fourth, comparing results across heterogeneous analyses requires a common metric. We transformed reported estimates into a standardized effect size (Cohen's $d$). Such transformations rest on assumptions that do not hold equally across all analytical settings \cite{Gelman.2013, Holzmeister.2024, Coretta.2023, vanAssen.2023}. Cohen's $d$ thus offers comparability at the cost of some precision. Finally, automated reanalysis assesses whether a result can be recovered from the original data. However, it does not test whether a finding holds in newly collected data. Reproducibility and replication are complementary. Confirming that a result is computationally reproducible does not establish that the underlying effect generalizes, which is why replication with newly collected data continues to remain relevant \cite{Zwaan.2018,Brodeur.2023}.


In sum, our findings suggest that LLMs can offer a path toward more scalable reproducibility assessment by automatically performing systematic reanalyses that are otherwise costly to conduct manually. Given the current capabilities and limitations of LLMs, such automated reanalysis should be viewed as a scalable screening tool rather than a substitute for expert judgment. Nevertheless, it could support the research community in quality control and provide a first-pass check for scientific journals, which may ultimately help improve rigor and reproducibility in empirical research.

\section*{Methods}

We preregistered our methods at \url{https://osf.io/84ue7} on June 10, 2026, prior to performing the reproducibility analysis. Before preregistration, we developed and tested the LLM pipeline on a small subset of 10 papers (see details below). 

\subsection*{Study sample}

We evaluated the LLM-based reproducibility pipeline on published empirical studies from the Systematizing Confidence in Open Research and Evidence (SCORE) initiative (\url{https://www.cos.io/score}) \cite{Alipourfard.2021}. SCORE is a stratified sample of articles published between 2009 and 2018 across major social and behavioral science disciplines, including psychology, economics, and political science. We filtered the original stratified sample of over 3000 papers for those studies for which original outcomes were provided by the SCORE project. We then kept only those studies for which a central empirical claim could be linked to a reported statistical result that was convertible to Cohen's $d$ (251 studies; Supplementary~Table~\ref{supptab:study_filtering}). Next, we excluded studies from both groups for which no data were available from the replication repositories (57 studies) or data were labeled as private (4 studies). To compute effect sizes, we extracted the reported effect size or a convertible statistic (\eg, $t$, $F$, $r$, $z$, $\chi^2$), the sample size, and the metadata needed to derive Cohen's $d$. We then mapped the statistics to Cohen's $d$ following the conversion procedure from \cite{Aczel.2026,multi100_conversion_code}, and excluded studies for which this conversion was not possible from the record alone (7 studies). One study was excluded due to a large Cohen's $d$ that was interpreted as a data error. 

We annotated the studies following the protocol in \cite{Aczel.2026} to obtain empirical claims phrased at the conceptual level. For each study, trained social and behavioral scientists wrote a short, standalone empirical claim in plain language that could be mapped to the corresponding inferential results from the paper. The goal was to capture the substantive relationship tested in the paper at the conceptual level, rather than to reproduce the statistical formulation or method-specific wording from the article. To do so, annotators used the existing SCORE information, which identified candidate claims and the corresponding statistical evidence, but often phrased these claims in statistical language or in wording close to the original paper, which included methodological details, model specifications, or other study-specific terminology. Each annotated claim had to (i)~be understandable on its own, (ii)~contain only one empirical relationship and a clear direction of effect, (iii)~correspond to a result based on a hypothesis test, and (iv)~be phrased at the conceptual rather than statistical level. For example, annotators aimed to capture claims such as whether one condition increased an outcome relative to another condition, or whether a predictor was positively or negatively associated with an outcome, without encoding the full statistical specification in the claim itself. Annotators were allowed to use the full article to support accurate interpretation of the claim and statistical result \cite{Aczel.2026}. Studies with no directional claim were excluded (2 studies). For the subset from \cite{Aczel.2026}, we reused the existing claim annotations and original effect sizes. 

Overall, the result is a standardized dataset of annotated empirical claims with matched statistical evidence, against which the LLM reanalysis can then be evaluated (see our code repository). Prior to preregistration, we used 10 studies to develop, test, and validate the analysis pipeline, including prompt design, code execution, quality control, and evaluation metrics. Excluding these studies left the main findings unchanged; i.e., the qualitative conclusion match rate was 79\% instead of 80\%, and the effect size was recovered for 23\% instead of 24\%.

A subset of these studies ($N=84$) had previously been analyzed in a large-scale human reanalysis effort \cite{Aczel.2026}. For this subset, we used the existing human reanalysis results as an additional benchmark for RQ2. This allowed us to compare the LLM-generated reanalyses not only against the original published findings, but also against independent human reanalyses of the same studies.

\subsection*{Data preparation}

For each study, we assembled a standardized study package consisting of (1)~the statistical claim, (2)~the original study data used in the publication, (3)~paper-specific metadata, and (4)~the article text. We converted article PDF files to Markdown using the Mistral OCR API (Mistral OCR~3 via the \texttt{mistral-ocr-latest} endpoint; model identifier \texttt{mistral-ocr-2512}, see \url{https://docs.mistral.ai/models/model-cards/ocr-3-25-12}). This step performs multimodal document parsing by combining optical character recognition (OCR) with layout-aware document extraction, which preserves the document structure even for complex scientific documents including their multi-column layouts, mathematical expressions, and tables. The extracted Markdown syntax thus captures the document structure, including section headings, paragraphs, and equations. Tables are extracted as Markdown-based representations, and figures as raster images.

\subsection*{LLM analysis pipeline}

\textbf{Agent task.} We implemented the analysis pipeline as an LLM agent that, for each study, receives the claim, the original dataset, and the paper, and is asked to write and execute statistical code in an isolated sandbox to test the claim. The agent does not have access to the original analysis code of the paper; it writes and executes code independently. Sandboxes are created fresh for each run, with no state carried over between runs, so that files, installed packages, and intermediate results from one run cannot influence subsequent runs. The agent submits a structured report containing the computed test statistic, sample size, degrees of freedom, and a qualitative conclusion (i.e., support / opposite / inconclusive). Each study is analyzed across 5 independent runs.

The prompt was designed following best practices for LLM prompt engineering \cite{Lin.2024,Giray.2023,Feuerriegel.2025}. The prompt casts the agent as a statistical analyst and asks it to commit to one reasonable analysis of the focal claim. The agent is explicitly instructed not to copy any test statistic reported in the paper; the submitted statistic must come from code the agent executes on the provided data. The full prompt is provided in Appendix~\ref{appendix:prompt}. 

In the supplementary materials, we additionally test the sensitivity to the framing of the prompt by comparing two alternatives: (1)~a ``confirmatory'' prompt, which instructed the agent to approach the analysis as a supportive reviewer expecting the claim to hold up under reanalysis, and (2)~a ``critical'' prompt, which instructed the agent to approach the analysis as a skeptical reviewer questioning whether the claim is as robust as the authors assert. Details are in Appendix~\ref{appendix:prompt}.

In the main analysis, the LLM agent receives the complete paper converted to Markdown, including the methods section. This design mirrors previous large-scale human reanalysis studies, in which the human reanalysts also had access to the full paper \cite{Aczel.2026}
. To assess whether the amount of methodological information affects automated reproducibility, we additionally conduct a sensitivity analysis where we vary the information context (see RQ3). For this analysis, we compare the default setting with the full paper as input against: (1)~a condition in which the methods section is removed from the Markdown file, and (2)~a condition in which the model receives only the abstract of the article. For the former, we manually removed methods-related sections from each paper, where relevant. These may include one or more sections titled, for example, ``Methods'', ``Data and methods'', ``Methodology'', ``Empirical approach'', ``Framework'', ``Empirical design'', or ``Research design'' (for five papers, no such section could be identified and instead relevant paragraphs with methodological specifications were removed). Figures and tables that are central to the removed methods sections, or referenced only within those sections, are also removed. In contrast, figures or tables that were positioned in the methods section were retained if they are substantively referenced elsewhere, such as in the results section. Appendices with supplementary methods were also removed. 

\textbf{Models.} We use Claude Opus~4.7 as the primary model. We further assess robustness with two additional frontier models, OpenAI GPT-5.5 (\texttt{openai/gpt-5.5}) and the open-weight Zhipu GLM-5.1 (\texttt{z-ai/glm-5.1}), so that our central findings do not depend on any single system; GLM-5.1 additionally tests whether the findings generalize beyond proprietary models. For all three models, we set the native reasoning effort to \emph{medium} (which we matched across models for comparability) and left all other decoding parameters at their provider defaults. We set a sampling temperature of $1.0$ for GPT-5.5 and GLM-5.1; Claude Opus~4.7 runs in always-on adaptive thinking and disregards the temperature parameter \cite{anthropic_adaptive_thinking}. Each LLM analysis run was capped at 100 messages, including reasoning, tool use, and submission steps, or a total API cost of US\$5, whichever was reached first. See Supplementary~Fig.~\ref{suppfig:model_comparison} for a comparison.

\textbf{Output standardization.} The agent was instructed to submit the final response as a structured JSON object. The output included the statistic type, test statistic, degrees of freedom, sample size, $p$-value, substantive conclusion (support / opposite / inconclusive), a short description of how the reanalysis was operationalized (including the dependent variable, main predictor, sample definition, model specification, controls, and a rationale), and the LLM reasoning trace.

We used the test statistic to compute a standardized effect size. Following \cite{Aczel.2026,multi100_conversion_code}, test statistics were first converted to a Pearson correlation coefficient and then transformed into Cohen's $d$. The conversion supported the statistic types requested from the agent: $t$, $F$, $z$, $\chi^2$, and $r$. When the required fields were missing or the statistic could not be converted, Cohen's $d$ was treated as unavailable for that run.

Following the preregistration, standardized effect sizes with $|d_{\mathrm{LLM}}| > 10$ were manually reviewed to determine the source of the result (e.g., whether this reflected a conversion or coding error rather than a plausible statistical estimate). Across all models and conditions, 20 of the 6,300 runs (0.3\%) exceeded this threshold. In 9 of the 20 runs, this occurred when the agent ran a nonparametric Wilcoxon signed-rank test but reported the rank-sum statistic $W=15$ under the $z$-statistic, thus processing $15$ as a $z$-score with $n=5$, which inflated Cohen's $d$ to $\approx 4\times10^{4}$. The remaining 11 cases were isolated runs in which the conversion failed (i.e., in 9 cases, the regression $t$-statistics were converted with a single (numerator) rather than the residual degree-of-freedom, and 2 cases reported an implausibly large $z \approx 31$). Because every affected study retained other valid runs, no study was dropped on this basis alone. Eventually, the 20 runs were excluded.

\textbf{Implementation details.} We implemented our LLM agent using the Inspect AI framework, which executes a ReAct-style loop of reasoning, tool use, and observation \cite{UK_AI_Security_Institute_Inspect_AI_2024}. The agent has access to four tools: (i)~a Python interpreter, (ii)~a bash shell, (iii)~ a \texttt{think} tool that allows the agent to record intermediate reasoning, and (iv)~a \texttt{view\_image} tool. Both the Python and bash tools return only text, while the \texttt{view\_image} tool allows the agent to visually inspect image files in the sandbox (used for reading figures that were extracted from papers). Note that GLM-5.1 is a text-only model (GLM-5.1); here, \texttt{view\_image} is disabled, and the agent relies on the parsed text and tables instead. The Python environment contains common scientific computing and statistical libraries; the complete list can be found in the system prompt (Appendix~\ref{appendix:prompt}).

\subsection*{Statistical analysis}

Following \cite{Aczel.2026}, all original effect sizes are reported as positive values, whereas the LLM and human reanalysis effect sizes are negative only when an ``opposite'' substantive conclusion is reported.

For each study, the 5 independent runs were aggregated into one effect size ($\bar{d}_{\mathrm{LLM}}$) using the mean. The qualitative conclusion is aggregated by taking the majority vote across the 5 runs; ties are resolved in favor of `inconclusive'. Studies were excluded if none of the 5 runs produced a valid effect size.

For each comparison, we compute $\Delta d = \bar{d}_{\mathrm{LLM}} - d_{\mathrm{original}}$, where $d_{\mathrm{original}}$ is the published effect size of the paper (or the mean human Cohen's $d$ for the comparison against human reanalysts). We classify a study as within-tolerance if $|\Delta d| \leq 0.05$ (strict) or $|\Delta d| \leq 0.20$ (broad). The primary metric is the proportion of studies within the strict tolerance \cite{Aczel.2026}.

\subsection*{Memorization test}

We conducted a separate memorization test \cite{Nori.2023, Sainz.2023} to assess whether model performance could be influenced by prior knowledge of the analyzed paper. We adapted this test to our setting by probing whether the LLM was able to recall the published effect size. For this, the agent received the paper reference and the focal claim as input, but not the original dataset, and was asked to report the expected effect size for the claim. Studies for which the response fell within $\pm 0.05$ Cohen's $d$ of the published result are flagged as potentially contaminated. Nevertheless, the results were largely robust (Supplementary~Table~\ref{supptab:memorization_test}).

\vspace{0.4cm}
\section*{Data availability}

Data from the Systematizing Confidence in Open Research and Evidence (SCORE) project \cite{Alipourfard.2021} are available at: \url{https://osf.io/dtzx4/overview}. The human reanalysis data are available from \cite{Aczel.2026}. Our GitHub repository (see Code Availability) contains our additional annotations and all derived analysis files needed to reproduce the results.

\vspace{0.4cm}
\section*{Code availability}

All code to replicate our analyses is available via GitHub at \url{https://github.com/tobihol/agentic-reproducibility}. The repository also contains the GUIDE-LLM checklist \cite{llm_checklist_2026} to document the LLM use.

\newpage
\bibliography{literature}

\begin{thebibliography}{10}
\expandafter\ifx\csname url\endcsname\relax
  \def\url#1{\texttt{#1}}\fi
\expandafter\ifx\csname urlprefix\endcsname\relax\def\urlprefix{URL }\fi
\providecommand{\bibinfo}[2]{#2}
\providecommand{\eprint}[2][]{\url{#2}}

\bibitem{Miske.2026}
\bibinfo{author}{Miske, O.} \emph{et~al.}
\newblock \bibinfo{title}{Investigating the reproducibility of the social and behavioural sciences}.
\newblock \emph{\bibinfo{journal}{Nature}} \textbf{\bibinfo{volume}{652}}, \bibinfo{pages}{126--134} (\bibinfo{year}{2026}).

\bibitem{Aczel.2026}
\bibinfo{author}{Aczel, B.} \emph{et~al.}
\newblock \bibinfo{title}{Investigating the analytical robustness of the social and behavioural sciences}.
\newblock \emph{\bibinfo{journal}{Nature}} \textbf{\bibinfo{volume}{652}}, \bibinfo{pages}{135--142} (\bibinfo{year}{2026}).

\bibitem{Brodeur.2026}
\bibinfo{author}{Brodeur, A.} \emph{et~al.}
\newblock \bibinfo{title}{Reproducibility and robustness of economics and political science research}.
\newblock \emph{\bibinfo{journal}{Nature}} \textbf{\bibinfo{volume}{652}}, \bibinfo{pages}{151--156} (\bibinfo{year}{2026}).

\bibitem{Hardwicke.2021}
\bibinfo{author}{Hardwicke, T.~E.} \emph{et~al.}
\newblock \bibinfo{title}{Analytic reproducibility in articles receiving open data badges at the journal \textit{{Psychological} {Science}}: An observational study}.
\newblock \emph{\bibinfo{journal}{Royal Society Open Science}} \textbf{\bibinfo{volume}{8}}, \bibinfo{pages}{201494} (\bibinfo{year}{2021}).

\bibitem{Brodeur.2024}
\bibinfo{author}{Brodeur, A.}, \bibinfo{author}{Mikola, D.} \& \bibinfo{author}{Cook, N.}
\newblock \bibinfo{title}{Mass reproducibility and replicability: A new hope}.
\newblock \bibinfo{type}{Tech. Rep.}, \bibinfo{institution}{SSRN} (\bibinfo{year}{2024}).

\bibitem{Fisar.2024}
\bibinfo{author}{Fišar, M.} \emph{et~al.}
\newblock \bibinfo{title}{Reproducibility in \textit{{Management} {Science}}}.
\newblock \emph{\bibinfo{journal}{Management Science}} \textbf{\bibinfo{volume}{70}}, \bibinfo{pages}{1343--1356} (\bibinfo{year}{2024}).

\bibitem{Nosek.2026}
\bibinfo{author}{Nosek, B.} \emph{et~al.}
\newblock \bibinfo{title}{Reimagining and diversifying assessment of the credibility of research findings}  (\bibinfo{year}{2026}).

\bibitem{Tyner.2026}
\bibinfo{author}{Tyner, A.~H.} \emph{et~al.}
\newblock \bibinfo{title}{Investigating the replicability of the social and behavioural sciences}.
\newblock \emph{\bibinfo{journal}{Nature}} \textbf{\bibinfo{volume}{652}}, \bibinfo{pages}{143--150} (\bibinfo{year}{2026}).

\bibitem{Parsons.2022}
\bibinfo{author}{Parsons, S.} \emph{et~al.}
\newblock \bibinfo{title}{A community-sourced glossary of open scholarship terms}.
\newblock \emph{\bibinfo{journal}{Nature Human Behaviour}} \textbf{\bibinfo{volume}{6}}, \bibinfo{pages}{312--318} (\bibinfo{year}{2022}).

\bibitem{OSC.2015}
\bibinfo{author}{{Open Science Collaboration}}.
\newblock \bibinfo{title}{Estimating the reproducibility of psychological science}.
\newblock \emph{\bibinfo{journal}{Science}} \textbf{\bibinfo{volume}{349}}, \bibinfo{pages}{aac4716} (\bibinfo{year}{2015}).

\bibitem{Peng.2011}
\bibinfo{author}{Peng, R.~D.}
\newblock \bibinfo{title}{Reproducible research in computational science}.
\newblock \emph{\bibinfo{journal}{Science}} \textbf{\bibinfo{volume}{334}}, \bibinfo{pages}{1226--1227} (\bibinfo{year}{2011}).

\bibitem{Sun.2026}
\bibinfo{author}{Sun, M.} \emph{et~al.}
\newblock \bibinfo{title}{{LAMBDA}: A large model based data agent}.
\newblock \emph{\bibinfo{journal}{Journal of the American Statistical Association}} \textbf{\bibinfo{volume}{121}}, \bibinfo{pages}{1--13} (\bibinfo{year}{2026}).

\bibitem{Boiko.2023}
\bibinfo{author}{Boiko, D.~A.}, \bibinfo{author}{MacKnight, R.}, \bibinfo{author}{Kline, B.} \& \bibinfo{author}{Gomes, G.}
\newblock \bibinfo{title}{Autonomous chemical research with large language models}.
\newblock \emph{\bibinfo{journal}{Nature}} \textbf{\bibinfo{volume}{624}}, \bibinfo{pages}{570--578} (\bibinfo{year}{2023}).

\bibitem{Qian.2024}
\bibinfo{author}{Qian, C.} \emph{et~al.}
\newblock \bibinfo{title}{{C}hat{D}ev: Communicative agents for software development}.
\newblock In \bibinfo{editor}{Ku, L.-W.}, \bibinfo{editor}{Martins, A.} \& \bibinfo{editor}{Srikumar, V.} (eds.) \emph{\bibinfo{booktitle}{Proceedings of the 62nd Annual Meeting of the Association for Computational Linguistics (Volume 1: Long Papers)}}, \bibinfo{pages}{15174--15186} (\bibinfo{publisher}{Association for Computational Linguistics}, \bibinfo{address}{Bangkok, Thailand}, \bibinfo{year}{2024}).

\bibitem{Lu.2026}
\bibinfo{author}{Lu, C.} \emph{et~al.}
\newblock \bibinfo{title}{Towards end-to-end automation of {AI} research}.
\newblock \emph{\bibinfo{journal}{Nature}} \textbf{\bibinfo{volume}{651}}, \bibinfo{pages}{914--919} (\bibinfo{year}{2026}).

\bibitem{Schmidgall.2025}
\bibinfo{author}{Schmidgall, S.} \& \bibinfo{author}{Moor, M.}
\newblock \bibinfo{title}{{AgentRxiv}: Towards collaborative autonomous research}.
\newblock \emph{\bibinfo{journal}{arXiv:2503.18102}}  (\bibinfo{year}{2025}).

\bibitem{Seo.2026}
\bibinfo{author}{Seo, M.}, \bibinfo{author}{Baek, J.}, \bibinfo{author}{Lee, S.} \& \bibinfo{author}{Hwang, S.~J.}
\newblock \bibinfo{title}{{Paper2Code}: Automating code generation from scientific papers in machine learning}.
\newblock \emph{\bibinfo{journal}{International Conference on Learning Representations (ICLR)}}  (\bibinfo{year}{2026}).

\bibitem{Alizadeh.2026}
\bibinfo{author}{Alizadeh, M.}, \bibinfo{author}{Mosleh, M.}, \bibinfo{author}{Gilardi, F.} \& \bibinfo{author}{Tucker, J.~A.}
\newblock \bibinfo{title}{Evaluating {AI} coding agents in social science reproducibility}  (\bibinfo{year}{2026}).

\bibitem{Kohler.2026}
\bibinfo{author}{Kohler, B.}, \bibinfo{author}{Zollikofer, D.}, \bibinfo{author}{Einsiedler, J.}, \bibinfo{author}{Hoyle, A.} \& \bibinfo{author}{Ash, E.}
\newblock \bibinfo{title}{Read the paper, write the code: Agentic reproduction of social-science results}.
\newblock \emph{\bibinfo{journal}{arXiv:2604.21965}}  (\bibinfo{year}{2026}).

\bibitem{Miao.2025}
\bibinfo{author}{Miao, J.}, \bibinfo{author}{Davis, J.~R.}, \bibinfo{author}{Zhang, Y.}, \bibinfo{author}{Pritchard, J.~K.} \& \bibinfo{author}{Zou, J.}
\newblock \bibinfo{title}{{Paper2Agent}: Reimagining research papers as interactive and reliable {AI} agents}.
\newblock \emph{\bibinfo{journal}{arXiv:2509.06917}}  (\bibinfo{year}{2025}).

\bibitem{Song.2025}
\bibinfo{author}{Song, Z.} \emph{et~al.}
\newblock \bibinfo{title}{Evaluating large language models in scientific discovery}.
\newblock \emph{\bibinfo{journal}{arXiv:2512.15567}}  (\bibinfo{year}{2025}).

\bibitem{Shao.2026}
\bibinfo{author}{Shao, E.} \emph{et~al.}
\newblock \bibinfo{title}{{SciSciGPT}: Advancing human--{AI} collaboration in the science of science}.
\newblock \emph{\bibinfo{journal}{Nature Computational Science}} \textbf{\bibinfo{volume}{6}}, \bibinfo{pages}{301--315} (\bibinfo{year}{2026}).

\bibitem{Zhang.2025}
\bibinfo{author}{Zhang, S.}, \bibinfo{author}{Fan, J.}, \bibinfo{author}{Fan, M.}, \bibinfo{author}{Li, G.} \& \bibinfo{author}{Du, X.}
\newblock \bibinfo{title}{{DeepAnalyze}: Agentic large language models for autonomous data science}.
\newblock \emph{\bibinfo{journal}{arXiv:2510.16872}}  (\bibinfo{year}{2025}).

\bibitem{Gottweis.2026}
\bibinfo{author}{Gottweis, J.} \emph{et~al.}
\newblock \bibinfo{title}{Accelerating scientific discovery with co-scientist}.
\newblock \emph{\bibinfo{journal}{Nature}}  (\bibinfo{year}{2026}).

\bibitem{Ghareeb.2026}
\bibinfo{author}{Ghareeb, A.~E.} \emph{et~al.}
\newblock \bibinfo{title}{A multi-agent system for automating scientific discovery}.
\newblock \emph{\bibinfo{journal}{Nature}} \bibinfo{pages}{forthcoming} (\bibinfo{year}{2026}).

\bibitem{Yamada.2025}
\bibinfo{author}{Yamada, Y.} \emph{et~al.}
\newblock \bibinfo{title}{The {AI} {Scientist-v2}: Workshop-level automated scientific discovery via agentic tree search}.
\newblock \emph{\bibinfo{journal}{arXiv:2504.08066}}  (\bibinfo{year}{2025}).

\bibitem{Song.2026}
\bibinfo{author}{Song, X.} \emph{et~al.}
\newblock \bibinfo{title}{{StatLLM}: A dataset for evaluating the performance of large language models in statistical analysis}.
\newblock \emph{\bibinfo{journal}{Scientific Data}} \textbf{\bibinfo{volume}{13}}, \bibinfo{pages}{369} (\bibinfo{year}{2026}).

\bibitem{Alipourfard.2021}
\bibinfo{author}{Alipourfard, N.} \emph{et~al.}
\newblock \bibinfo{title}{Systematizing confidence in open research and evidence (score)} (\bibinfo{year}{2021}).
\newblock \urlprefix\url{osf.io/preprints/socarxiv/46mnb_v1}.

\bibitem{Silberzahn.2018}
\bibinfo{author}{Silberzahn, R.} \emph{et~al.}
\newblock \bibinfo{title}{Many analysts, one data set: Making transparent how variations in analytic choices affect results}.
\newblock \emph{\bibinfo{journal}{Advances in Methods and Practices in Psychological Science}} \textbf{\bibinfo{volume}{1}}, \bibinfo{pages}{337--356} (\bibinfo{year}{2018}).

\bibitem{BotvinikNezer.2020}
\bibinfo{author}{Botvinik-Nezer, R.} \emph{et~al.}
\newblock \bibinfo{title}{Variability in the analysis of a single neuroimaging dataset by many teams}.
\newblock \emph{\bibinfo{journal}{Nature}} \textbf{\bibinfo{volume}{582}}, \bibinfo{pages}{84--88} (\bibinfo{year}{2020}).

\bibitem{Steegen.2016}
\bibinfo{author}{Steegen, S.}, \bibinfo{author}{Tuerlinckx, F.}, \bibinfo{author}{Gelman, A.} \& \bibinfo{author}{Vanpaemel, W.}
\newblock \bibinfo{title}{Increasing transparency through a multiverse analysis}.
\newblock \emph{\bibinfo{journal}{Perspectives on Psychological Science}} \textbf{\bibinfo{volume}{11}}, \bibinfo{pages}{702--712} (\bibinfo{year}{2016}).

\bibitem{Breznau.2022}
\bibinfo{author}{Breznau, N.} \emph{et~al.}
\newblock \bibinfo{title}{Observing many researchers using the same data and hypothesis reveals a hidden universe of uncertainty}.
\newblock \emph{\bibinfo{journal}{Proceedings of the National Academy of Sciences}} \textbf{\bibinfo{volume}{119}}, \bibinfo{pages}{e2203150119} (\bibinfo{year}{2022}).

\bibitem{Auspurg.2021}
\bibinfo{author}{Auspurg, K.} \& \bibinfo{author}{Br{\"u}derl, J.}
\newblock \bibinfo{title}{Has the credibility of the social sciences been credibly destroyed? {R}eanalyzing the “many analysts, one data set” project}.
\newblock \emph{\bibinfo{journal}{Socius}} \textbf{\bibinfo{volume}{7}}, \bibinfo{pages}{23780231211024421} (\bibinfo{year}{2021}).

\bibitem{Scheel.2022}
\bibinfo{author}{Scheel, A.~M.}
\newblock \bibinfo{title}{Why most psychological research findings are not even wrong}.
\newblock \emph{\bibinfo{journal}{Infant and Child Development}} \textbf{\bibinfo{volume}{31}}, \bibinfo{pages}{e2295} (\bibinfo{year}{2022}).

\bibitem{Oberauer.2019}
\bibinfo{author}{Oberauer, K.} \& \bibinfo{author}{Lewandowsky, S.}
\newblock \bibinfo{title}{Addressing the theory crisis in psychology}.
\newblock \emph{\bibinfo{journal}{Psychonomic Bulletin \& Review}} \textbf{\bibinfo{volume}{26}}, \bibinfo{pages}{1596--1618} (\bibinfo{year}{2019}).

\bibitem{Coretta.2023}
\bibinfo{author}{Coretta, S.} \emph{et~al.}
\newblock \bibinfo{title}{Multidimensional signals and analytic flexibility: Estimating degrees of freedom in human-speech analyses}.
\newblock \emph{\bibinfo{journal}{Advances in Methods and Practices in Psychological Science}} \textbf{\bibinfo{volume}{6}}, \bibinfo{pages}{25152459231162567} (\bibinfo{year}{2023}).

\bibitem{Gelman.2013}
\bibinfo{author}{Gelman, A.} \& \bibinfo{author}{Loken, E.}
\newblock \bibinfo{title}{The garden of forking paths: Why multiple comparisons can be a problem, even when there is no ``fishing expedition'' or ``$p$-hacking'' and the research hypothesis was posited ahead of time}.
\newblock \bibinfo{type}{Tech. Rep.}, \bibinfo{institution}{Department of Statistics, Columbia University}, \bibinfo{address}{New York, NY} (\bibinfo{year}{2013}).

\bibitem{Patel.2015}
\bibinfo{author}{Patel, C.~J.}, \bibinfo{author}{Burford, B.} \& \bibinfo{author}{Ioannidis, J.~P.}
\newblock \bibinfo{title}{Assessment of vibration of effects due to model specification can demonstrate the instability of observational associations}.
\newblock \emph{\bibinfo{journal}{Journal of Clinical Epidemiology}} \textbf{\bibinfo{volume}{68}}, \bibinfo{pages}{1046--1058} (\bibinfo{year}{2015}).

\bibitem{Wagenmakers.2022}
\bibinfo{author}{Wagenmakers, E.-J.}, \bibinfo{author}{Sarafoglou, A.} \& \bibinfo{author}{Aczel, B.}
\newblock \bibinfo{title}{One statistical analysis must not rule them all}.
\newblock \emph{\bibinfo{journal}{Nature}} \textbf{\bibinfo{volume}{605}}, \bibinfo{pages}{423--425} (\bibinfo{year}{2022}).

\bibitem{Nuijten.2020}
\bibinfo{author}{Nuijten, M.~B.} \& \bibinfo{author}{Polanin, J.~R.}
\newblock \bibinfo{title}{“statcheck”: Automatically detect statistical reporting inconsistencies to increase reproducibility of meta-analyses}.
\newblock \emph{\bibinfo{journal}{Research Synthesis Methods}} \textbf{\bibinfo{volume}{11}}, \bibinfo{pages}{574--579} (\bibinfo{year}{2020}).

\bibitem{Bertran.2026}
\bibinfo{author}{Bertran, M.}, \bibinfo{author}{Fogliato, R.} \& \bibinfo{author}{Wu, Z.~S.}
\newblock \bibinfo{title}{Many {AI} analysts, one dataset: Navigating the agentic data science multiverse}.
\newblock \emph{\bibinfo{journal}{arXiv:2602.18710}}  (\bibinfo{year}{2026}).

\bibitem{Brodeur.2025}
\bibinfo{author}{Brodeur, A.} \emph{et~al.}
\newblock \bibinfo{title}{Comparing human-only, {AI}-assisted, and {AI}-led teams on assessing research reproducibility in quantitative social science}.
\newblock \bibinfo{type}{IZA Discussion Paper} \bibinfo{number}{17645}, \bibinfo{institution}{Institute of Labor Economics (IZA)}, \bibinfo{address}{Bonn} (\bibinfo{year}{2025}).

\bibitem{Siegel.2024}
\bibinfo{author}{Siegel, Z.~S.}, \bibinfo{author}{Kapoor, S.}, \bibinfo{author}{Nagdir, N.}, \bibinfo{author}{Stroebl, B.} \& \bibinfo{author}{Narayanan, A.}
\newblock \bibinfo{title}{{CORE-Bench}: Fostering the credibility of published research through a computational reproducibility agent benchmark}.
\newblock \emph{\bibinfo{journal}{Transactions on Machine Learning Research}}  (\bibinfo{year}{2024}).

\bibitem{Starace.2025}
\bibinfo{author}{Starace, G.} \emph{et~al.}
\newblock \bibinfo{title}{{PaperBench}: Evaluating {AI}'s ability to replicate {AI} research}.
\newblock In \emph{\bibinfo{booktitle}{International Conference on Machine Learning}}, \bibinfo{pages}{56843--56873} (\bibinfo{year}{2025}).

\bibitem{Wrightson.2025}
\bibinfo{author}{Wrightson, J.~G.}, \bibinfo{author}{Blazey, P.}, \bibinfo{author}{Moher, D.}, \bibinfo{author}{Khan, K.~M.} \& \bibinfo{author}{Ardern, C.~L.}
\newblock \bibinfo{title}{{GPT} for {RCTs}? {U}sing {AI} to determine adherence to clinical trial reporting guidelines}.
\newblock \emph{\bibinfo{journal}{BMJ Open}} \textbf{\bibinfo{volume}{15}}, \bibinfo{pages}{e088735} (\bibinfo{year}{2025}).

\bibitem{Nosek.2018}
\bibinfo{author}{Nosek, B.~A.}, \bibinfo{author}{Ebersole, C.~R.}, \bibinfo{author}{DeHaven, A.~C.} \& \bibinfo{author}{Mellor, D.~T.}
\newblock \bibinfo{title}{The preregistration revolution}.
\newblock \emph{\bibinfo{journal}{Proceedings of the National Academy of Sciences}} \textbf{\bibinfo{volume}{115}}, \bibinfo{pages}{2600--2606} (\bibinfo{year}{2018}).

\bibitem{Brodeur.2020}
\bibinfo{author}{Brodeur, A.}, \bibinfo{author}{Cook, N.} \& \bibinfo{author}{Heyes, A.}
\newblock \bibinfo{title}{Methods matter: $p$-hacking and publication bias in causal analysis in economics}.
\newblock \emph{\bibinfo{journal}{American Economic Review}} \textbf{\bibinfo{volume}{110}}, \bibinfo{pages}{3634--3660} (\bibinfo{year}{2020}).

\bibitem{Simmons.2011}
\bibinfo{author}{Simmons, J.~P.}, \bibinfo{author}{Nelson, L.~D.} \& \bibinfo{author}{Simonsohn, U.}
\newblock \bibinfo{title}{False-positive psychology: Undisclosed flexibility in data collection and analysis allows presenting anything as significant}.
\newblock \emph{\bibinfo{journal}{Psychological Science}} \textbf{\bibinfo{volume}{22}}, \bibinfo{pages}{1359--1366} (\bibinfo{year}{2011}).

\bibitem{Nori.2023}
\bibinfo{author}{Nori, H.}, \bibinfo{author}{King, N.}, \bibinfo{author}{McKinney, S.~M.}, \bibinfo{author}{Carignan, D.} \& \bibinfo{author}{Horvitz, E.}
\newblock \bibinfo{title}{Capabilities of {GPT-4} on medical challenge problems}.
\newblock \emph{\bibinfo{journal}{arXiv:2303.13375}}  (\bibinfo{year}{2023}).

\bibitem{Sainz.2023}
\bibinfo{author}{Sainz, O.} \emph{et~al.}
\newblock \bibinfo{title}{{NLP} evaluation in trouble: On the need to measure {LLM} data contamination for each benchmark}.
\newblock In \emph{\bibinfo{booktitle}{Findings of the Association for Computational Linguistics: EMNLP 2023}}, \bibinfo{pages}{10776–10787} (\bibinfo{publisher}{Association for Computational Linguistics}, \bibinfo{year}{2023}).

\bibitem{Krahmer.2026}
\bibinfo{author}{Krähmer, D.}, \bibinfo{author}{Schächtele, L.} \& \bibinfo{author}{Auspurg, K.}
\newblock \bibinfo{title}{Code sharing and reproducibility in survey-based social research: Evidence from a large-scale audit}.
\newblock \emph{\bibinfo{journal}{Royal Society Open Science}} \textbf{\bibinfo{volume}{13}}, \bibinfo{pages}{251997} (\bibinfo{year}{2026}).

\bibitem{Holzmeister.2024}
\bibinfo{author}{Holzmeister, F.} \emph{et~al.}
\newblock \bibinfo{title}{Heterogeneity in effect size estimates}.
\newblock \emph{\bibinfo{journal}{Proceedings of the National Academy of Sciences}} \textbf{\bibinfo{volume}{121}}, \bibinfo{pages}{e2403490121} (\bibinfo{year}{2024}).

\bibitem{vanAssen.2023}
\bibinfo{author}{van Assen, M.~A.}, \bibinfo{author}{Stoevenbelt, A.~H.} \& \bibinfo{author}{van Aert, R.~C.}
\newblock \bibinfo{title}{The end justifies all means: Questionable conversion of different effect sizes to a common effect size measure}.
\newblock \emph{\bibinfo{journal}{Religion, Brain \& Behavior}} \textbf{\bibinfo{volume}{13}}, \bibinfo{pages}{345--347} (\bibinfo{year}{2023}).

\bibitem{Zwaan.2018}
\bibinfo{author}{Zwaan, R.~A.}, \bibinfo{author}{Etz, A.}, \bibinfo{author}{Lucas, R.~E.} \& \bibinfo{author}{Donnellan, M.~B.}
\newblock \bibinfo{title}{Making replication mainstream}.
\newblock \emph{\bibinfo{journal}{Behavioral and Brain Sciences}} \textbf{\bibinfo{volume}{41}}, \bibinfo{pages}{e120} (\bibinfo{year}{2018}).

\bibitem{Brodeur.2023}
\bibinfo{author}{Brodeur, A.}, \bibinfo{author}{Dreber, A.}, \bibinfo{author}{Hoces de~la Guardia, F.} \& \bibinfo{author}{Miguel, E.}
\newblock \bibinfo{title}{Replication games: How to make reproducibility research more systematic}.
\newblock \emph{\bibinfo{journal}{Nature}} \textbf{\bibinfo{volume}{621}}, \bibinfo{pages}{684--686} (\bibinfo{year}{2023}).

\bibitem{multi100_conversion_code}
\bibinfo{title}{{Multi100 conversion code}}.
\newblock \bibinfo{howpublished}{\url{https://github.com/marton-balazs-kovacs/multi100/blob/47c0b8c6dd68e19eb80fa8843dce18f0d3655ae1/analysis/multi100_raw_processed.qmd\#L160-L165}}.

\bibitem{Lin.2024}
\bibinfo{author}{Lin, Z.}
\newblock \bibinfo{title}{How to write effective prompts for large language models}.
\newblock \emph{\bibinfo{journal}{Nature Human Behaviour}} \textbf{\bibinfo{volume}{8}}, \bibinfo{pages}{611–615} (\bibinfo{year}{2024}).

\bibitem{Giray.2023}
\bibinfo{author}{Giray, L.}
\newblock \bibinfo{title}{Prompt engineering with {ChatGPT}: A guide for academic writers}.
\newblock \emph{\bibinfo{journal}{Annals of Biomedical Engineering}} \textbf{\bibinfo{volume}{51}}, \bibinfo{pages}{2629–2633} (\bibinfo{year}{2023}).

\bibitem{Feuerriegel.2025}
\bibinfo{author}{Feuerriegel, S.} \emph{et~al.}
\newblock \bibinfo{title}{Using natural language processing to analyse text data in behavioural science}.
\newblock \emph{\bibinfo{journal}{Nature Reviews Psychology}} \textbf{\bibinfo{volume}{4}}, \bibinfo{pages}{96--111} (\bibinfo{year}{2025}).

\bibitem{anthropic_adaptive_thinking}
\bibinfo{author}{{Anthropic}}.
\newblock \bibinfo{title}{Adaptive thinking}.
\newblock \bibinfo{howpublished}{Claude API Documentation} (\bibinfo{year}{2026}).
\newblock \urlprefix\url{https://platform.claude.com/docs/en/build-with-claude/adaptive-thinking}.
\newblock \bibinfo{note}{Accessed: 2026-06-22}.

\bibitem{UK_AI_Security_Institute_Inspect_AI_2024}
\bibinfo{author}{{AI Security Institute, UK}}.
\newblock \bibinfo{title}{Inspect {AI}: {Framework} for {Large} {Language} {Model} {Evaluations}}.
\newblock \bibinfo{howpublished}{\url{https://github.com/UKGovernmentBEIS/inspect_ai}} (\bibinfo{year}{2024}).
\newblock \bibinfo{note}{Software}.

\bibitem{llm_checklist_2026}
\bibinfo{author}{Feuerriegel, S.} \emph{et~al.}
\newblock \bibinfo{title}{A reporting checklist for large language models in behavioural science}.
\newblock \emph{\bibinfo{journal}{Nature Human Behaviour}}  (\bibinfo{year}{2026}).

\end{thebibliography}


\newpage
\section*{Acknowledgments}

Funding from the Deutsche Forschungsgemeinschaft (DFG, German Research Foundation) under the National Research Data Infrastructure – NFDI 27/1-2026, project number 460037581 is acknowledged. SF acknowledges funding via the Swiss National Science Foundation (SNSF), Grant 186932. Our research is supported by the DAAD programme Konrad Zuse Schools of Excellence in Artificial Intelligence, sponsored by the Federal Ministry of Research, Technology and Space.

\vspace{0.4cm}
\section*{Author contributions} 

All authors contributed to conceptualization, manuscript writing, and approved the manuscript.

\vspace{0.4cm}
\section*{Competing interests}
The authors declare no competing interests.

\newpage

\appendix

\section*{Appendix}

\setcounter{figure}{0}
\setcounter{table}{0}

\renewcommand{\thefigure}{S\arabic{figure}}
\renewcommand{\thetable}{S\arabic{table}}

\renewcommand{\figurename}{Fig.}
\renewcommand{\tablename}{Table}

\section{Study sample}

\begin{table}[H]
\centering
\singlespacing
\caption{\textbf{Study filtering and inclusion.} Counts show the number of studies retained and excluded when constructing the evaluation corpus.}
\label{supptab:study_filtering}
\begin{tabular}{lrrr}
\toprule
\textbf{Filtering step} & \textbf{Multi100} \cite{Aczel.2026} & \textbf{SCORE extension} & \textbf{Total} \\
\midrule
Candidate studies & 100 & 151 & 251 \\
\midrule
Excluded: no data available & 0 & 40 & 40 \\
Excluded: no data in repository data node & 7 & 10 & 17 \\
Excluded: private data & 4 & 0 & 4 \\
Excluded: Cohen's $d$ not computable & 5$^\dagger$ & 2 & 7 \\
Excluded: outlier Cohen's $d$ & 0 & 1 & 1 \\
Excluded: no directional claim & 0 & 2 & 2 \\
\midrule
Total excluded & 16 & 55 & 71 \\
\midrule
Included in evaluation corpus & 84 & 96 & 180 \\
\bottomrule
\multicolumn{3}{p{.8\textwidth}}{$^\dagger$ This is consistent with the original Multi100 \cite{Aczel.2026} study in which original effect sizes could not be determined due to missing information.}
\end{tabular}
\end{table}

\newpage

\section{Robustness across different LLMs (strict tolerance)}


\begin{figure}
\centering
\includegraphics[width=0.95\linewidth]{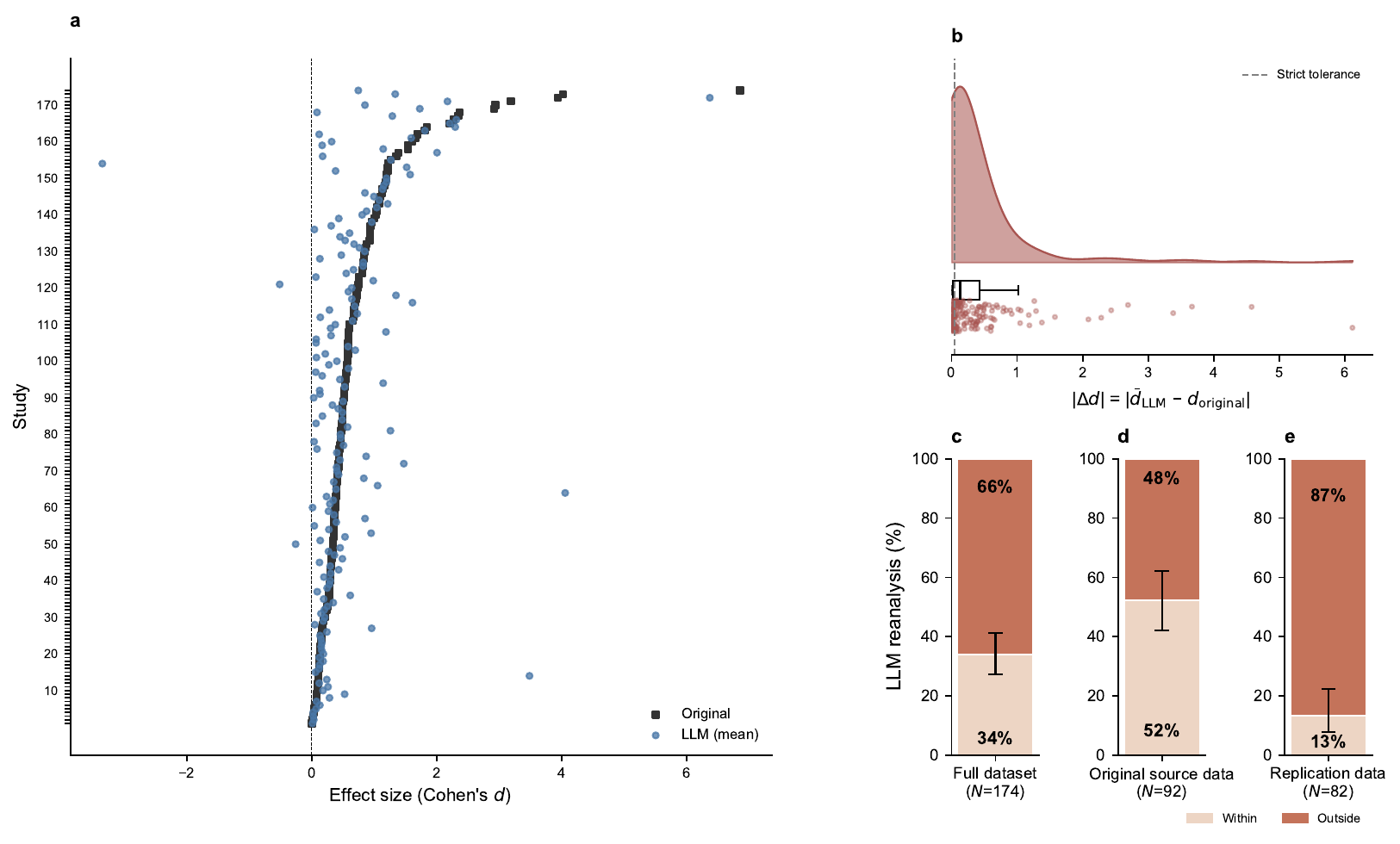}
\caption{\textbf{Automated reproducibility assessment using GPT-5.5.} 
\textbf{a,}~Effect size of the original analysis (gray squares; all represented as positive values) and the effect sizes of the reanalyses (blue dots) for each study. 
\textbf{b,}~Distribution of $|\Delta d|$, computed as $|\Delta d| = |\bar{d}_{\mathrm{LLM}} - d_{\mathrm{original}}|$.
The distribution is visualized using a density plot, a boxplot, and scatter points. In the boxplot, the line indicates the median, the box denotes the interquartile range (IQR), and the whiskers extend to 1.5$\times$ of the IQR; points beyond the whiskers indicate outliers.
\textbf{c,}~Proportion of studies for which the LLM-generated effect size falls within or outside the tolerance region around the original result, across the full sample. \textbf{d,}~Proportion of studies falling within or outside the tolerance region for the subset of studies that make the original source data available. \textbf{e,}~Proportion of studies falling within or outside the tolerance region for the subset of studies where the original source data were unavailable but where replication data for reanalyses are provided. For 6 studies, the LLM did not produce a valid Cohen's $d$, and the resulting studies were excluded, leaving $N=174$.
Whiskers indicate Wilson 95\% confidence intervals.
}
\label{suppfig:rq1_main_result_gpt}
\end{figure}

\begin{figure}
\centering
\includegraphics[width=0.95\linewidth]{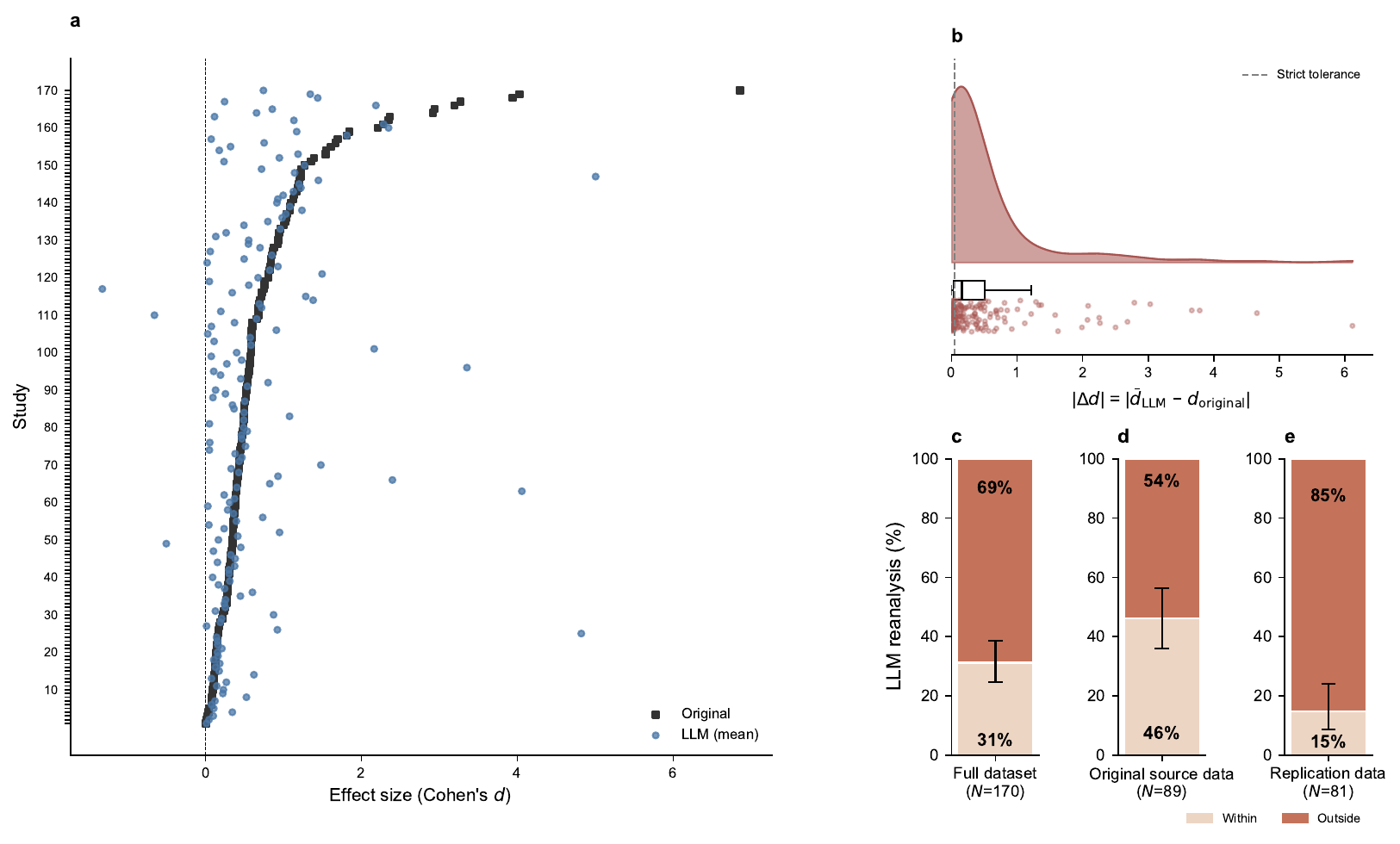}
\caption{\textbf{Automated reproducibility assessment using GLM-5.1.} 
\textbf{a,}~Effect size of the original analysis (gray squares; all represented as positive values) and the effect sizes of the reanalyses (blue dots) for each study. 
\textbf{b,}~Distribution of $|\Delta d|$, computed as $|\Delta d| = |\bar{d}_{\mathrm{LLM}} - d_{\mathrm{original}}|$.
The distribution is visualized using a density plot, a boxplot, and scatter points. In the boxplot, the line indicates the median, the box denotes the interquartile range (IQR), and the whiskers extend to 1.5$\times$ of the IQR; points beyond the whiskers indicate outliers.
\textbf{c,}~Proportion of studies for which the LLM-generated effect size falls within or outside the tolerance region around the original result, across the full sample. \textbf{d,}~Proportion of studies falling within or outside the tolerance region for the subset of studies that make the original source data available. \textbf{e,}~Proportion of studies falling within or outside the tolerance region for the subset of studies where the original source data were unavailable but where replication data for reanalyses are provided. For 10 studies, the LLM did not produce a valid Cohen's $d$, and the resulting studies were excluded, leaving $N=170$.
Whiskers indicate Wilson 95\% confidence intervals.
}
\label{suppfig:rq1_main_result_glm}
\end{figure}

\newpage

\section{Robustness across different LLMs (broad tolerance)}


\begin{figure}
\centering
\includegraphics[width=0.95\linewidth]{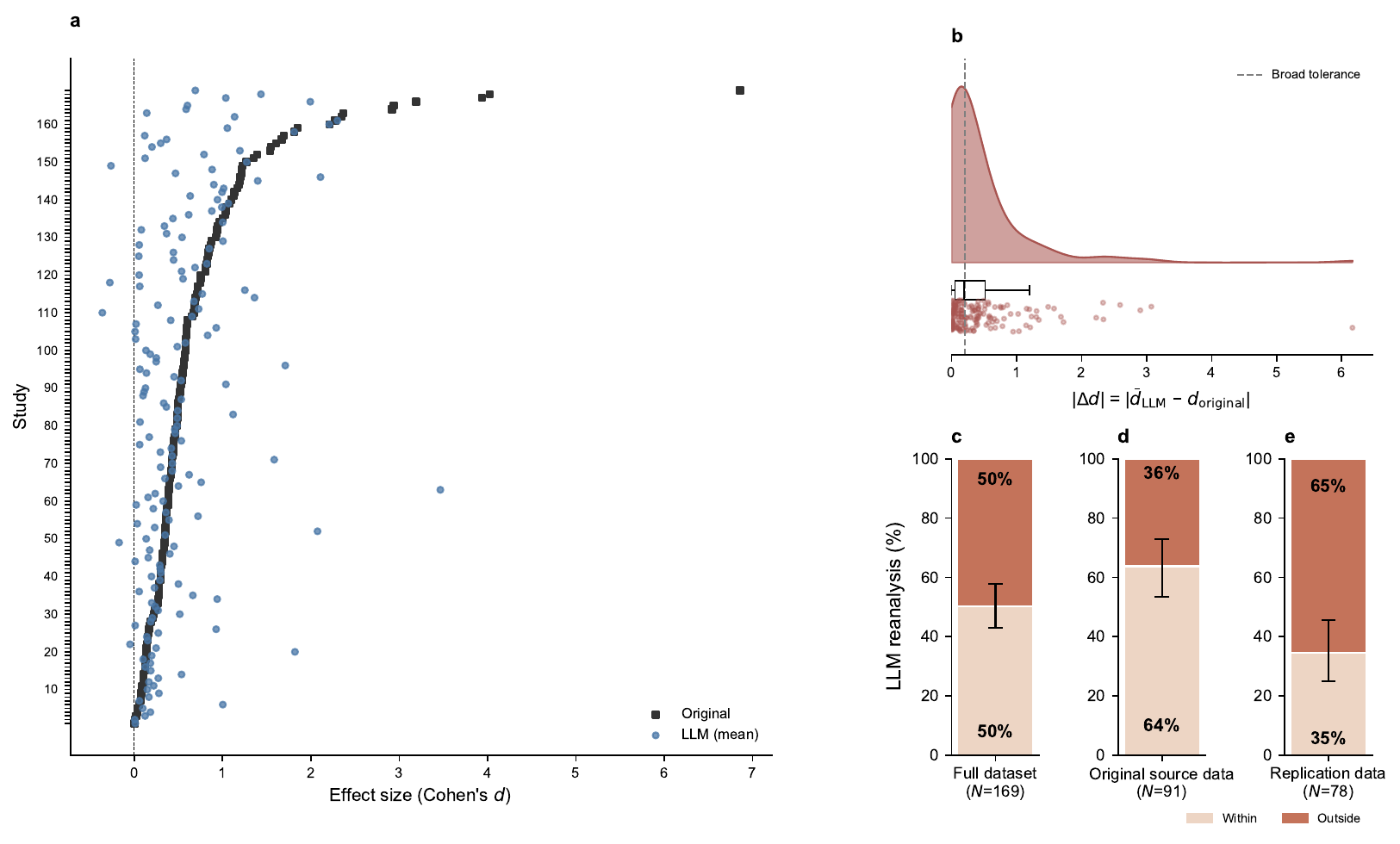}
\caption{\textbf{Automated reproducibility assessment using Claude Opus~4.7 with a broad tolerance ($\pm0.20$ Cohen's $d$).}
\textbf{a,}~Effect size of the original analysis (gray squares; all represented as positive values) and the effect sizes of the reanalyses (blue dots) for each study. 
\textbf{b,}~Distribution of $|\Delta d|$, computed as $|\Delta d| = |\bar{d}_{\mathrm{LLM}} - d_{\mathrm{original}}|$.
The distribution is visualized using a density plot, a boxplot, and scatter points. In the boxplot, the line indicates the median, the box denotes the interquartile range (IQR), and the whiskers extend to 1.5$\times$ of the IQR; points beyond the whiskers indicate outliers.
\textbf{c,}~Proportion of studies for which the LLM-generated effect size falls within or outside the tolerance region around the original result, across the full sample. \textbf{d,}~Proportion of studies falling within or outside the tolerance region for the subset of studies that make the original source data available. \textbf{e,}~Proportion of studies falling within or outside the tolerance region for the subset of studies where the original source data were unavailable but where replication data for reanalyses are provided. For 11 studies, the LLM did not produce a valid Cohen's $d$, and the resulting studies were excluded, leaving $N=169$.
Whiskers indicate Wilson 95\% confidence intervals. 
}
\label{suppfig:rq1_main_result_broad}
\end{figure}

\begin{figure}
\centering
\includegraphics[width=0.95\linewidth]{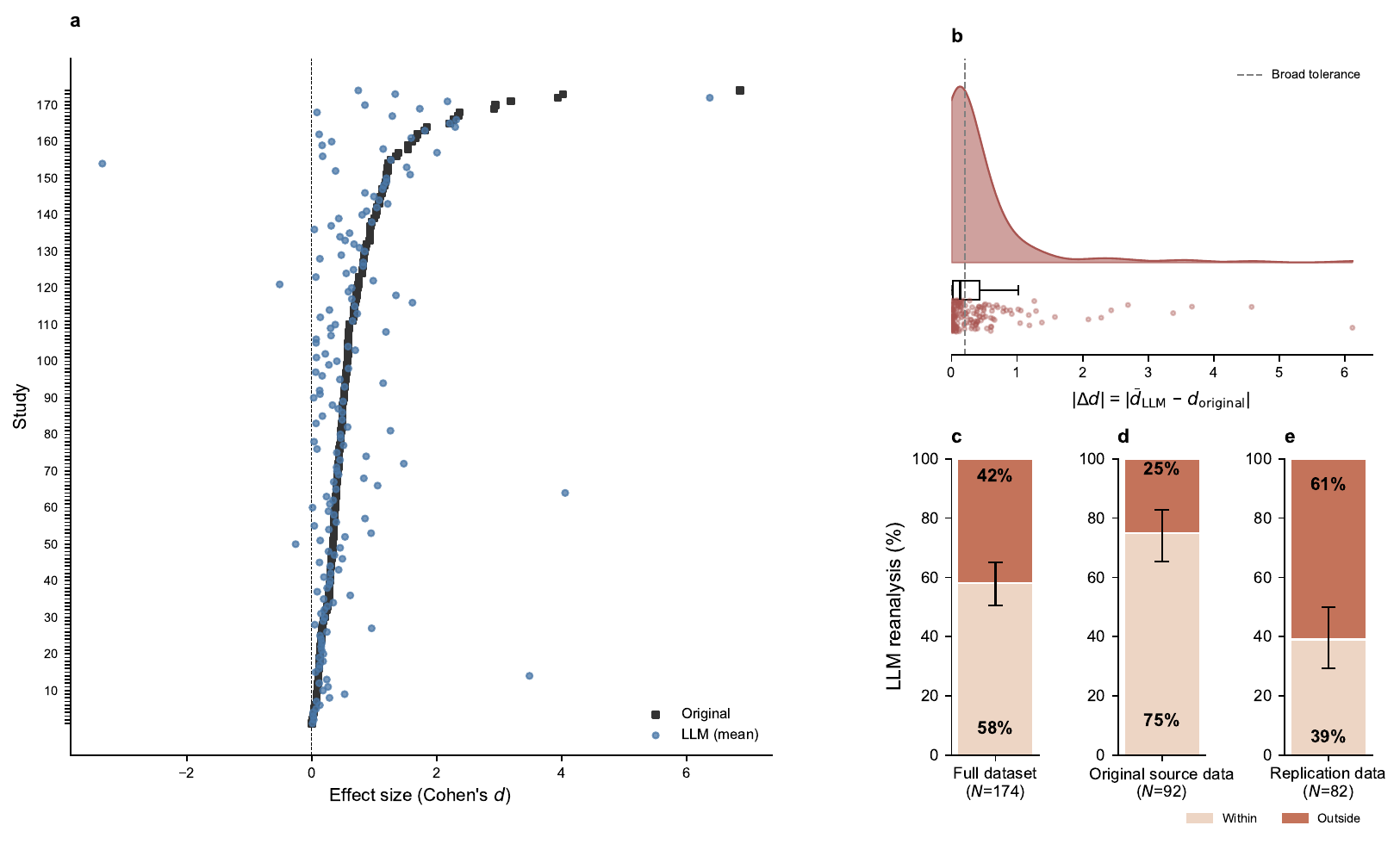}
\caption{\textbf{Automated reproducibility assessment using GPT-5.5 with a broad tolerance ($\pm0.20$ Cohen's $d$).}
\textbf{a,}~Effect size of the original analysis (gray squares; all represented as positive values) and the effect sizes of the reanalyses (blue dots) for each study. 
\textbf{b,}~Distribution of $|\Delta d|$, computed as $|\Delta d| = |\bar{d}_{\mathrm{LLM}} - d_{\mathrm{original}}|$.
The distribution is visualized using a density plot, a boxplot, and scatter points. In the boxplot, the line indicates the median, the box denotes the interquartile range (IQR), and the whiskers extend to 1.5$\times$ of the IQR; points beyond the whiskers indicate outliers.
\textbf{c,}~Proportion of studies for which the LLM-generated effect size falls within or outside the tolerance region around the original result, across the full sample. \textbf{d,}~Proportion of studies falling within or outside the tolerance region for the subset of studies that make the original source data available. \textbf{e,}~Proportion of studies falling within or outside the tolerance region for the subset of studies where the original source data were unavailable but where replication data for reanalyses are provided. For 6 studies, the LLM did not produce a valid Cohen's $d$, and the resulting studies were excluded, leaving $N=174$.
Whiskers indicate Wilson 95\% confidence intervals.
}
\label{suppfig:rq1_main_result_gpt_broad}
\end{figure}

\begin{figure}
\centering
\includegraphics[width=0.95\linewidth]{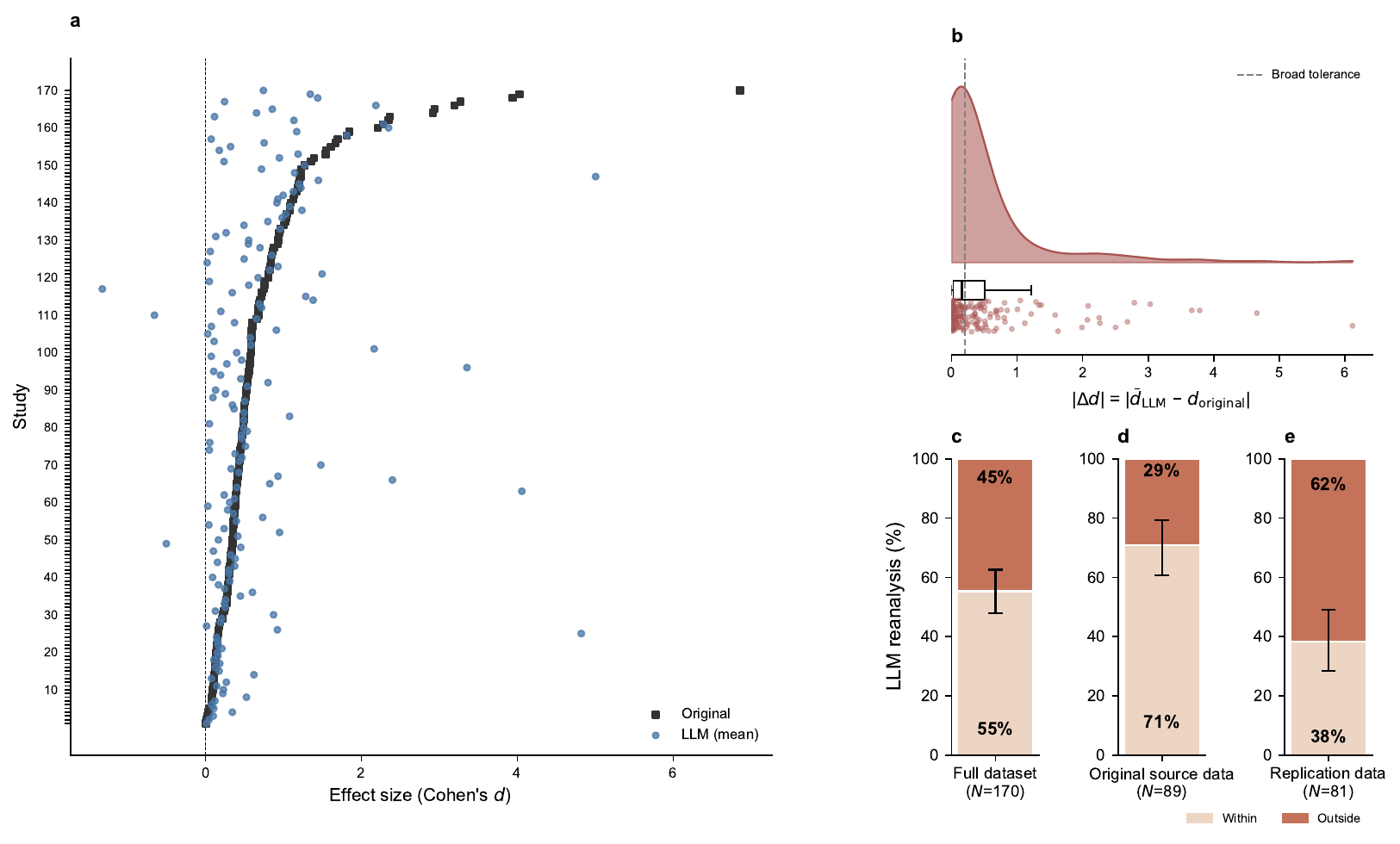}
\caption{\textbf{Automated reproducibility assessment using GLM-5.1 with a broad tolerance ($\pm0.20$ Cohen's $d$).}
\textbf{a,}~Effect size of the original analysis (gray squares; all represented as positive values) and the effect sizes of the reanalyses (blue dots) for each study. 
\textbf{b,}~Distribution of $|\Delta d|$, computed as $|\Delta d| = |\bar{d}_{\mathrm{LLM}} - d_{\mathrm{original}}|$.
The distribution is visualized using a density plot, a boxplot, and scatter points. In the boxplot, the line indicates the median, the box denotes the interquartile range (IQR), and the whiskers extend to 1.5$\times$ of the IQR; points beyond the whiskers indicate outliers.
\textbf{c,}~Proportion of studies for which the LLM-generated effect size falls within or outside the tolerance region around the original result, across the full sample. \textbf{d,}~Proportion of studies falling within or outside the tolerance region for the subset of studies that make the original source data available. \textbf{e,}~Proportion of studies falling within or outside the tolerance region for the subset of studies where the original source data were unavailable but where replication data for reanalyses are provided.
For 10 studies, the LLM did not produce a valid Cohen's $d$, and the resulting studies were excluded, leaving $N=170$.
Whiskers indicate Wilson 95\% confidence intervals.
}
\label{suppfig:rq1_main_result_glm_broad}
\end{figure}

\newpage

\section{Sensitivity to confirmatory vs. critical prompt framing}

We examined whether the analytical perspective induced by prompt framing influenced LLM reproducibility. We compared three conditions, all run the on full paper variant using Claude Opus~4.7: (i)~a \emph{neutral} baseline using the base system prompt; (ii)~a \emph{confirmatory} condition, in which an instruction to approach the analysis with a prior that the claim is empirically robust was appended to the system prompt; and (iii)~a \emph{critical} condition, in which an instruction to approach the analysis as a skeptical reviewer was appended. 

\begin{figure}
\centering
\includegraphics[width=0.95\linewidth]{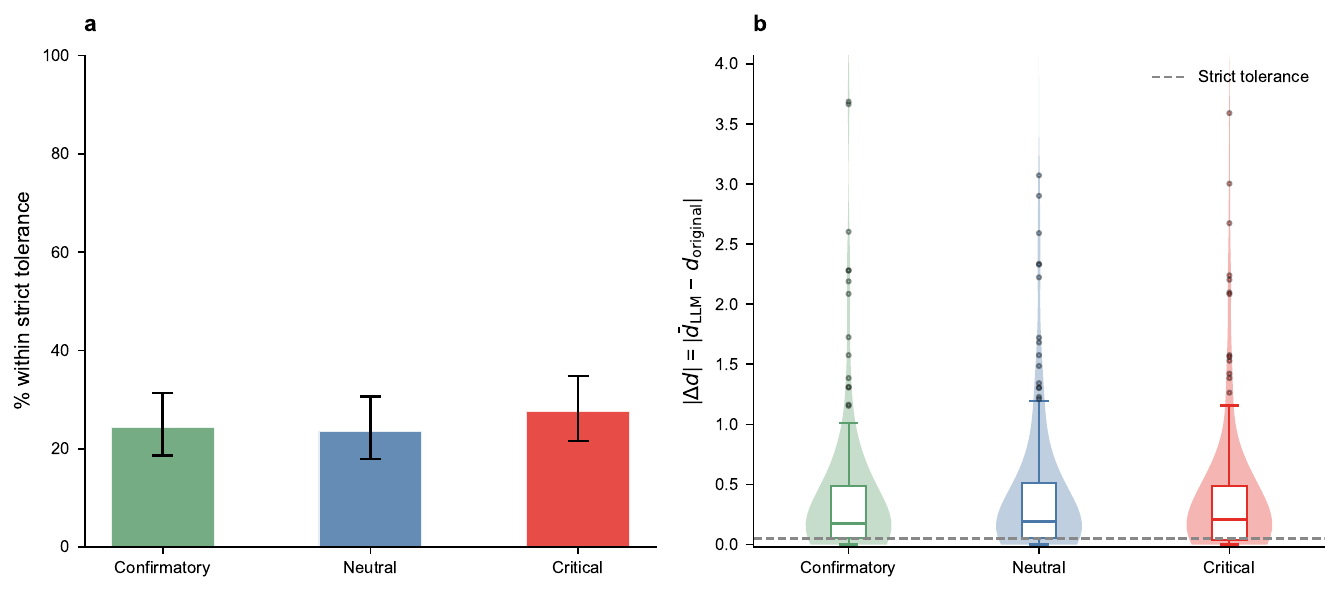}
\caption{\textbf{Sensitivity to confirmatory vs. critical prompt framing.}
\textbf{a,}~Proportion of studies where the mean LLM-generated effect size falls within $\pm$0.05 Cohen's $d$ of the original. The results are separately shown for the confirmatory, neutral, and critical prompt framing conditions.
Whiskers show Wilson 95\% confidence intervals. 
\textbf{b,}~Distribution of absolute effect size deviations (i.e., $|\Delta d| = |\bar{d}_{\text{LLM}} - d_{\text{original}}|$) by different framings. The dashed line shows the $\pm$0.05 tolerance threshold.
In the boxplot, the line indicates the median, the box denotes the interquartile range (IQR), and the whiskers extend to 1.5$\times$ of the IQR; points beyond the whiskers indicate outliers.
Results are based on Claude~Opus~4.7 over five independent runs per study.
}
\label{suppfig:perspective}
\end{figure}

Within the strict $\pm 0.05$ tolerance band, $\bar{d}_{\mathrm{LLM}}$ fell within tolerance in 24\% of studies for the neutral condition, 24\% for the confirmatory condition, and  28\% for the critical condition (Supplementary~Fig.~\ref{suppfig:perspective}). The conclusion (using a majority vote over the five independent runs) matched the original in 80\%, 82\%, and 77\% of studies for neutral, confirmatory, and critical conditions, respectively. These results provide no clear evidence that confirmatory or critical prompt framing systematically alters reproducibility relative to the neutral baseline. 

\section{Heterogeneity analysis}

\begin{figure}
\centering
\includegraphics[width=0.95\linewidth]{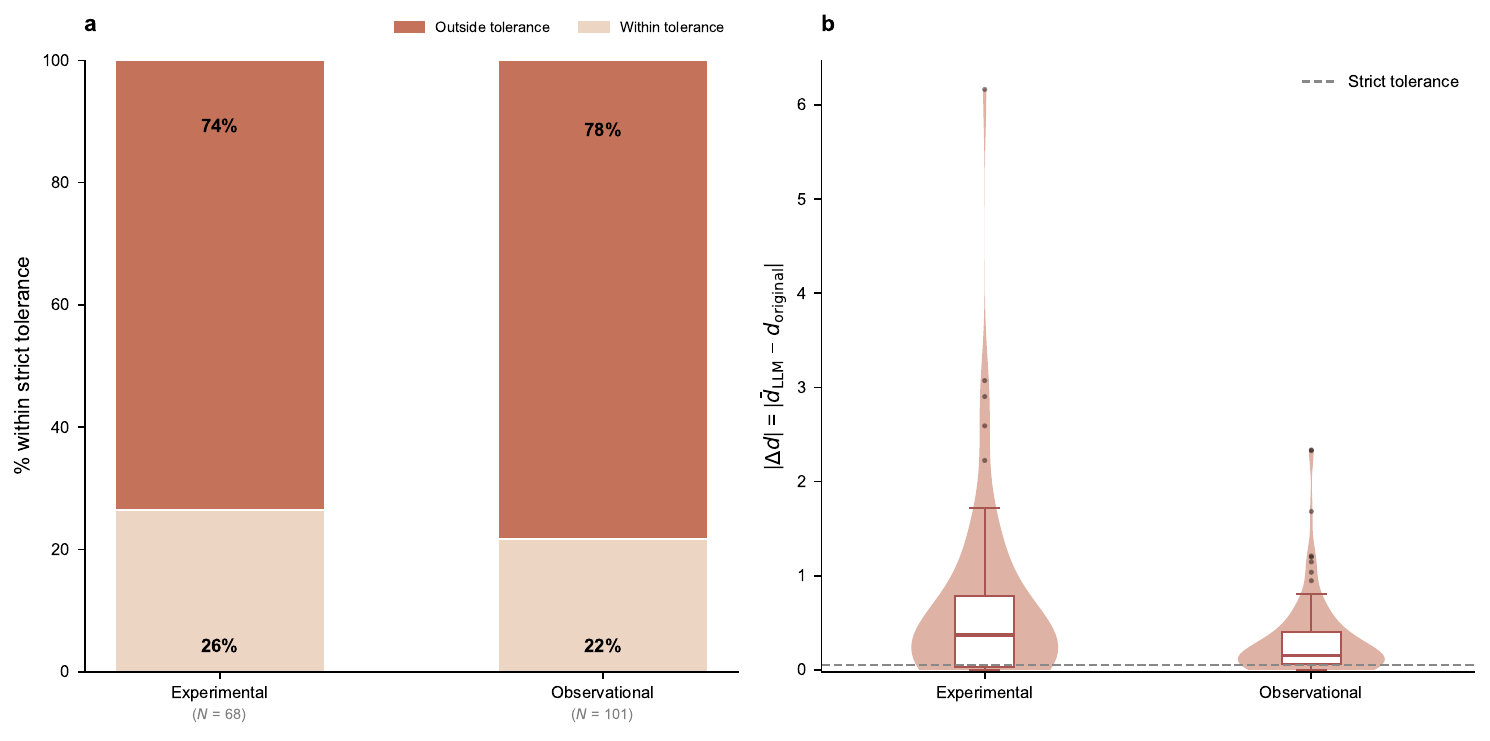}
\caption{\textbf{Reproducibility grouped by study type (experimental vs. observational).}
\textbf{a,}~Proportion of studies where the mean LLM-generated effect size falls within $\pm$0.05 Cohen's $d$ grouped by study type. 
\textbf{b,}~Distribution of absolute effect size deviations (i.e., $|\Delta d| = |\bar{d}_{\text{LLM}} - d_{\text{original}}|$) grouped by study type. 
In the boxplot, the line indicates the median, the box denotes the interquartile range (IQR), and the whiskers extend to 1.5$\times$ of the IQR; points beyond the whiskers indicate outliers.
Results are based on Claude~Opus~4.7 over five independent runs per study.
}
\label{suppfig:by_study_characteristics}
\end{figure}

\begin{figure}
\centering
\includegraphics[width=0.95\linewidth]{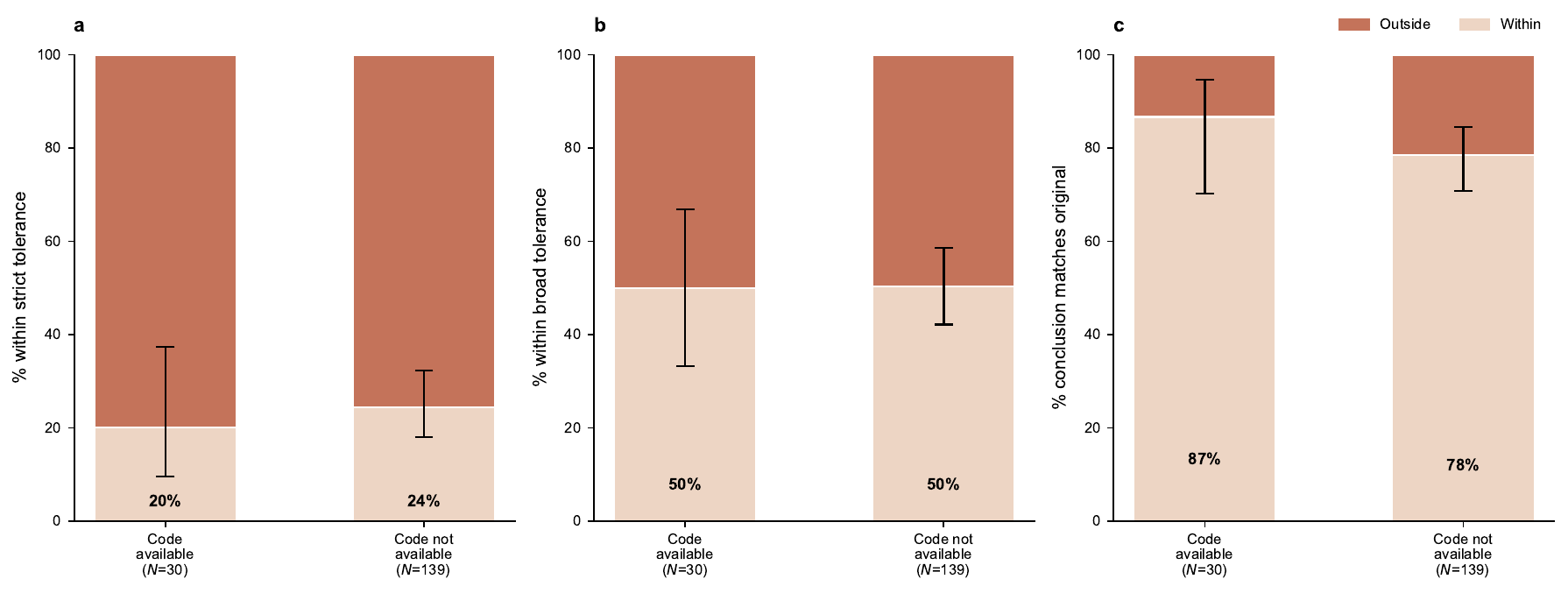}
\caption{\textbf{Heterogeneity by public code availability.}
Studies are grouped according to whether the original analysis code was publicly available. 
\textbf{a,}~Proportion of studies for which the mean generated effect size falls within the strict tolerance region ($\pm0.05$ Cohen's $d$) around the original effect size. 
\textbf{b,}~Proportion of studies for which the mean generated effect size falls within the broad tolerance region ($\pm0.20$ Cohen's $d$) around the original effect size. 
\textbf{c,}~Proportion of studies for which the majority-vote conclusion matches the original conclusion.
Whiskers indicate Wilson 95\% confidence intervals. 
Results are based on Claude Opus~4.7 over five independent runs per study.
}
\label{suppfig:code_availability}
\end{figure}

\newpage

\begin{figure}
\centering
\includegraphics[width=0.95\linewidth]{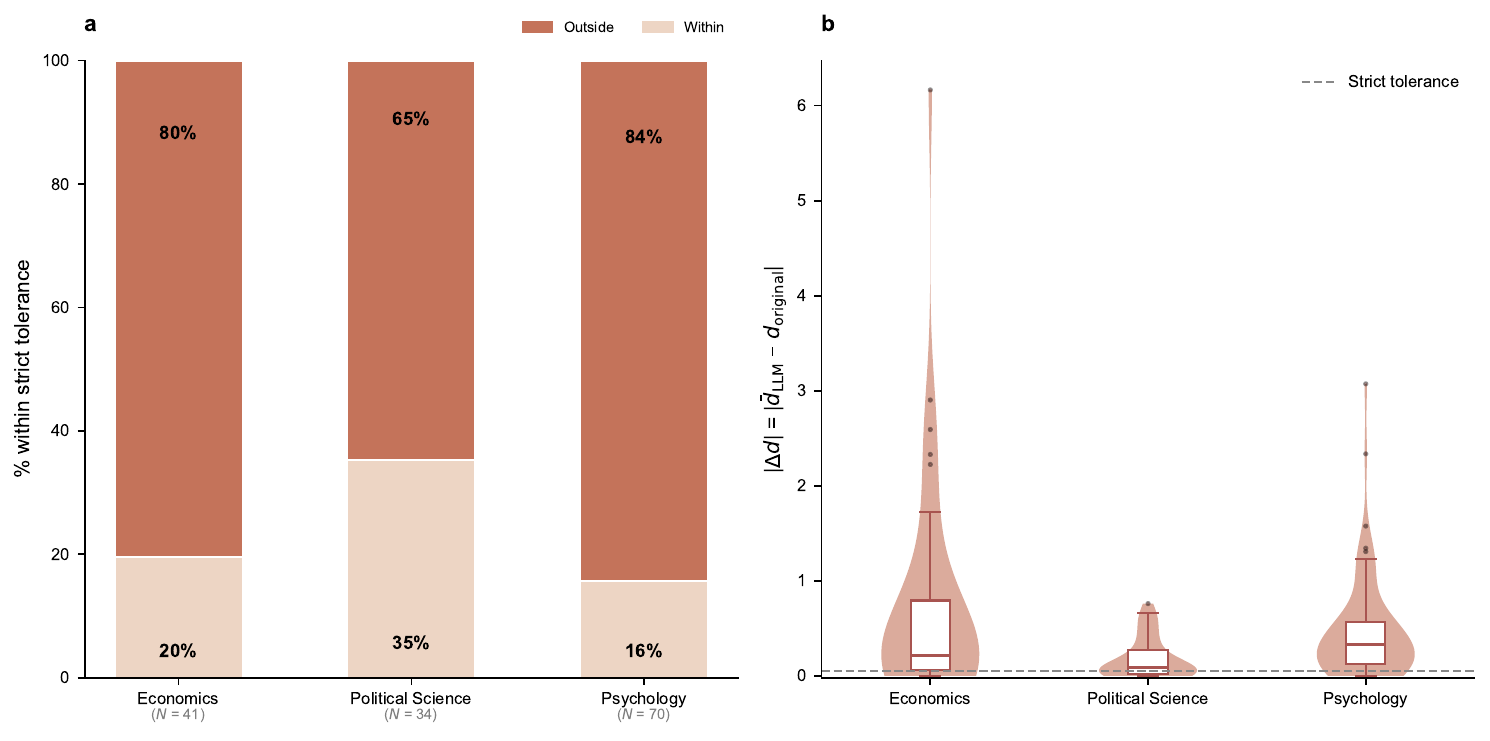}
\caption{\textbf{Heterogeneity by disciplines (economics, political science, psychology).}
\textbf{a,}~Proportion of studies where the mean LLM effect size falls within $\pm$0.05 Cohen's $d$ of the original effect size, grouped by discipline.
\textbf{b,}~Distribution of absolute effect size deviations (i.e., $|\Delta d| = |\bar{d}_{\text{LLM}} - d_{\text{original}}|$) grouped by discipline. The dashed line shows the $\pm0.05$ tolerance threshold. 
In the boxplot, the line indicates the median, the box denotes the interquartile range (IQR), and the whiskers extend to 1.5$\times$ of the IQR; points beyond the whiskers indicate outliers. Note that not all studies from the SCORE sample could be matched to the three disciplines.
Results are based on Claude~Opus~4.7 over five independent runs per study.
}
\label{suppfig:by_discipline}
\end{figure}

\newpage

\begin{figure}
\centering
\includegraphics[width=0.95\linewidth]{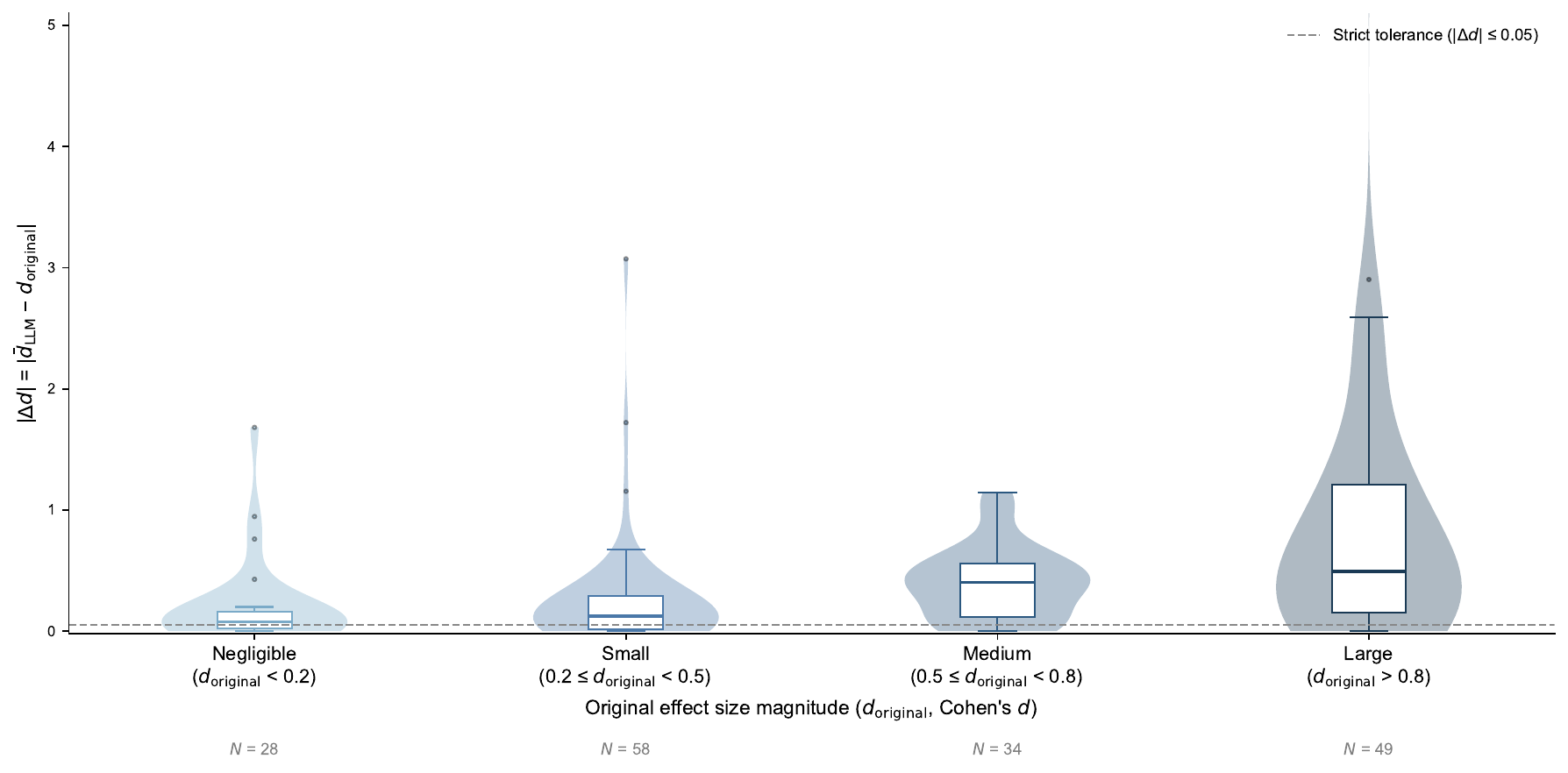}
\caption{\textbf{Reproducibility by original effect size.}
Studies are grouped by the magnitude of the original Cohen's $d$: negligible ($<0.2$), small ($[0.2, 0.5)$), medium ($[0.5, 0.8)$), and large ($\geq0.8$). This analysis assesses whether reproducibility varies with the size of the originally reported effect.
Under the strict criterion (i.e., $|\Delta d| \leq 0.05$), accuracy is highest for small ($33\%$) and negligible ($29\%$) original effects, but falls to $16\%$ for large and $15\%$ for medium effects.
In the boxplot, the line indicates the median, the box denotes the interquartile range (IQR), and the whiskers extend to 1.5$\times$ of the IQR; points beyond the whiskers indicate outliers.
Results are based on Claude~Opus~4.7 over five independent runs per study.
}
\label{suppfig:es_magnitude}
\end{figure}

\begin{figure}
\centering
\includegraphics[width=0.80\linewidth]{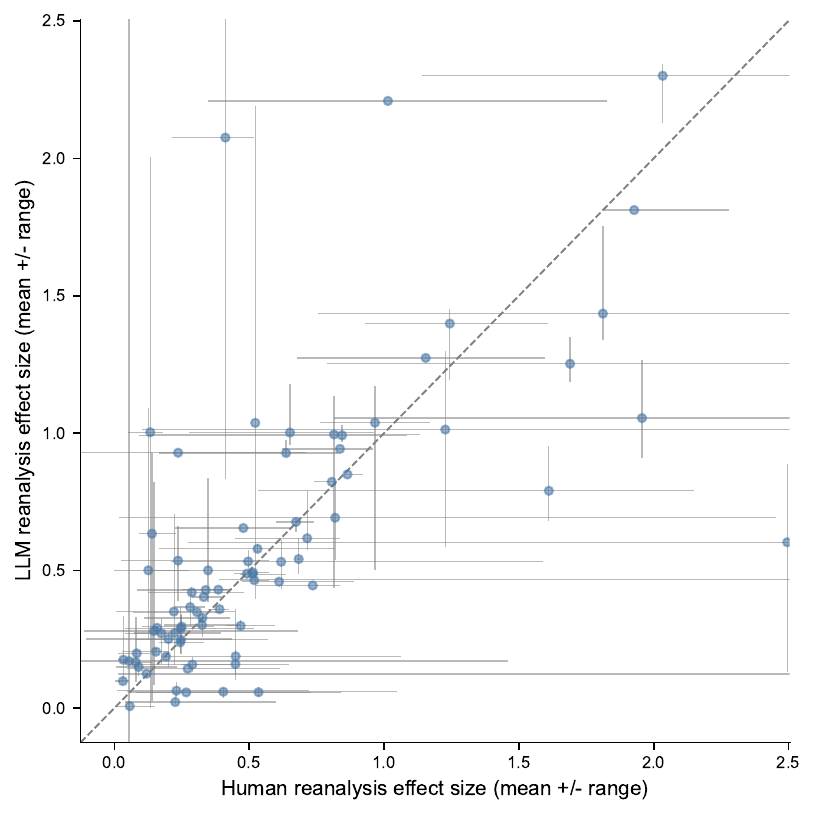}
\caption{\textbf{Analytical variability in LLM-generated and human reanalyses.}
Each point represents one study from the subset with human reanalysis benchmarks. The x-axis shows the mean Cohen's $d$ across human reanalyses, and the y-axis shows the mean generated Cohen's $d$ across five independent runs. Horizontal whiskers indicate the minimum-to-maximum range across human analysts; vertical whiskers indicate the minimum-to-maximum range across LLM runs. 
Results are based on Claude~Opus~4.7 over five independent runs per study.
}
\label{suppfig:effect_size_ranges}
\end{figure}

\begin{figure}
\centering
\includegraphics[width=0.80\linewidth]{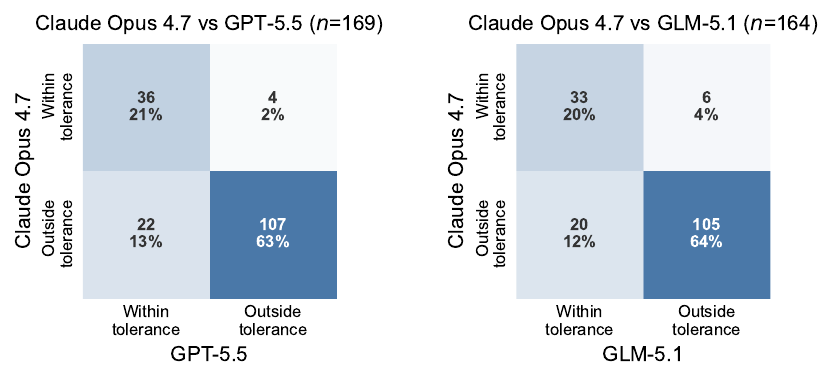}
\caption{\textbf{Cross-model agreement on effect-size within $\pm0.05$ Cohen's $d$ (strict tolerance).} Confusion matrices comparing whether Claude Opus~4.7 and a second model recover the original effect size on the same studies, for RQ1 (full paper text, neutral framing). \textbf{Left,}~Claude Opus~4.7 versus GPT-5.5 ($N = 169$). \textbf{Right,}~Claude Opus~4.7 versus GLM-5.1 ($N = 164$). Each $N$ is restricted to the studies for which both models produced a valid mean estimate. Cells report the number of studies and their percentage of $N$. Diagonal cells (both within / both outside) denote agreement, off-diagonal cells denote the studies where one model recovers the original effect and the other does not. Overall agreement is 85\% (Claude Opus~4.7 vs.\ GPT-5.5) and 84\% (Claude Opus~4.7 vs.\ GLM-5.1).}
\label{suppfig:model_comparison}
\end{figure}

\newpage

\section{Memorization test}

We conducted a memorization test to assess whether model performance could be influenced by prior exposure to the analyzed papers. The prompt focused on two dimensions. (1)~We asked whether the LLM recalled the paper, yielding a binary recall indicator. (2)~To probe whether such recall extended to the focal quantitative result, we asked the model to output the Cohen's $d$ for the focal claim and then checked whether the recalled value fell within the strict ($\pm0.05$) or broad ($\pm0.20$) tolerance region around the original effect size. Results and the exact prompt are reported below. 

\begin{table}[H]
\centering
\singlespacing
\footnotesize
\caption{\textbf{Memorization test.}
The memorization test assessed (1)~whether the model reported recalling the paper; and (2)~when probed for the focal quantitative result, whether the recalled Cohen's $d$ matched the original effect size within predefined tolerance regions. Although the model reported recalling 14 papers, none of the recalled Cohen's $d$ values fell within either the strict ($\pm0.05$) or broad ($\pm0.20$) tolerance region.}
\label{supptab:memorization_test}
\begin{tabular}{lr}
\toprule
\textbf{Measure} & \textbf{Result} \\
\midrule
Papers reported as recalled by the model & 14/180 \\
\midrule
Recalled Cohen's $d$ within $\pm0.05$ of the original effect size & 0\% \\
Recalled Cohen's $d$ within $\pm0.20$ of the original effect size & 0\% \\
\bottomrule
\multicolumn{1}{l}{\emph{Note:} Results are based on Claude~Opus~4.7.}
\end{tabular}
\end{table}

\vspace{0.35cm}

\begin{minipage}{0.9\linewidth}
\footnotesize
\textbf{Memorization test: prompt}
\begin{lstlisting}[style=promptstyle]
I am studying a published paper and want to know what you know about its findings.

## Paper Reference
{paper_reference}

## Research Claim Being Tested
{claim}

Please answer the following questions based solely on your training knowledge (do NOT make up or guess, only report what you actually know):

1. Do you have knowledge of this specific paper in your training data? (yes / uncertain / no)
2. If yes: What is the main finding regarding the claim above? Be specific.
3. If yes: What is the direction of the effect? (positive / negative / null / unknown)
4. If yes: Report the main test statistic in this structured format, type (z/t/F/chi2/r), numeric value, degrees of freedom (if applicable), and sample size.
5. How confident are you in your recall of this paper's results? (1-10, where 10 = certain)

Fill in the block below. Use "unknown" for any field you do not know.

```probe_results
PAPER_KNOWN: [yes / uncertain / no]
RECALLED_FINDING: [brief description of finding, or "unknown"]
RECALLED_DIRECTION: [positive / negative / null / unknown]
RECALLED_STAT_TYPE: [z / t / F / chi2 / r / unknown]
RECALLED_STAT_VALUE: [numeric value or unknown]
RECALLED_DF1: [numeric or null or unknown]
RECALLED_DF2: [numeric or null or unknown]
RECALLED_SAMPLE_SIZE: [integer or unknown]
RECALL_CONFIDENCE: [1-10]
```
\end{lstlisting}
\end{minipage}

\newpage
\section{Prompt}
\label{appendix:prompt}

This appendix documents the full prompt configuration used in all LLM analyses. The prompt has three components: a fixed \emph{system prompt} applied in every condition, a per-paper \emph{user prompt} assembled from a template, and a condition-specific \emph{perspective instruction} added to the system prompt only for the prompt framing analysis.

\medskip
\noindent\textbf{System prompt}

\begin{lstlisting}[style=promptstyle]
You are a statistical analyst. The user will give you a published study's claim, the original dataset, and (depending on the experimental condition) either the full original paper or a partial version of it. Your task is to compute the test statistic most directly tied to the target claim from the data yourself on the original data and report the numerical statistic in the requested structured format.

## Tools

You have `python`, `bash`, `think`, and `view_image`. The Python sandbox includes pandas, numpy, scipy, statsmodels, pyreadstat, pyreadr, xlrd, openpyxl, h5py, mat73, oct2py, pyarrow, duckdb, py7zr, and pymupdf4llm. Octave is installed system-wide (callable from oct2py).
**State is not preserved. ** Each call to `python` or `bash` runs in a fresh process - variables, imports, and loaded data do NOT persist between calls. Write complete, self-contained scripts each time, or save intermediate results to disk (e.g., CSV or pickle) and reload them in subsequent calls.

## Data formats

Installed readers cover: `.dta .sav .por .sas7bdat .xpt` (pyreadstat), `.csv .tsv .txt .dat .json .xlsx` (pandas/openpyxl), `.xls` (xlrd), `.RData .rda .rds` (pyreadr), `.mat` (scipy.io / mat73 / oct2py), `.parquet .feather` (pyarrow), `.h5 .hdf5` (h5py), `.duckdb .db` (duckdb / sqlite3), `.zip .tar.gz .7z` (unzip / tar / py7zr).
If the relevant data file cannot be read with these libraries, submit immediately with `test_statistic: null` and explain why - do not attempt byte-level parsing or manual reconstruction.

## Operationalization

A verbal claim like "X increases Y" rarely pins down a single analysis - there are multiple reasonable choices of subsample, dependent variable, model, and controls that all "test the claim." When the original paper is provided, use its methods section to guide your operationalization choices. When it is not, commit to one specific reasonable operationalization based on the data structure and the claim, and execute it.

## Computing the result yourself

The reported test statistic must come from the analysis you run on the data. If the paper happens to be visible to you and reports a test statistic for the claim, you must NOT copy that number. Your job is to evaluate whether the claim reproduces from the provided data using the analysis most directly tied to the claim. Run the analysis in the execution environment and report the statistic YOUR code produces.

## Submission

Do NOT call submit() until you have actually run the analysis and obtained real numerical results from the data. Do NOT guess, approximate, or fabricate values from the paper or your prior knowledge - if you cannot compute a particular field, report null in it and explain why in your reasoning. The user prompt will specify the exact submission schema.
\end{lstlisting}

\medskip
\noindent\textbf{User prompt}

The user prompt is assembled per paper from a fixed template. Placeholders in angle brackets (\texttt{<...>}) are filled with the paper's title, identifier, and claim, and with the sandbox paths to the dataset and parsed paper. The \emph{Materials} block is the only part that varies across the RQ3 information-level conditions. The full-paper version is shown inline, with the redacted methods and abstract-only variants reported below it.

\begin{lstlisting}[style=promptstyle]
You have been assigned to reanalyze the following published study:

*Paper Title:* <paper title>
*Paper ID:* <paper id>

*Claim:* <claim>

This claim asserts the presence of a directional effect or relationship in the data. Your task is to conduct an independent analysis on the original dataset to test it.

## Instructions

Conduct **one analysis** on the dataset to test the claim. The verbal claim does not uniquely determine the analysis - there will be more than one defensible choice of subsample, dependent variable, model, and controls. **Commit to one specific operationalization** and execute it. Do not enumerate alternatives or report multiple results.

Your analysis must produce a **single, main result** expressed as a **z-, t-, F-, or chi-squared test** (or its non-parametric equivalent), with the test statistic, degrees of freedom, and sample size needed to compute a standardised effect size. The reported statistic must come from your analysis on the data - not from any number printed in the paper. Draw a single substantive conclusion from your result.

## Materials

Data directory: `<data dir>`
Original paper: Markdown file `<paper path>`.

Use the paper's methods section to guide your operationalization (subsample, dependent variable, model specification). The paper may also report a test statistic for this claim - do not copy it. Run the analysis on the data yourself and report the statistic your code produces. Tables are available as readable Markdown files (`tbl-N.md`) in the same `<paper dir>/` directory. Figures referenced inline in the paper Markdown (e.g. `![img-N.jpeg](img-N.jpeg)`) are present in `<paper dir>/` - use the `view_image` tool to look at one.

## Reporting Results

When you are done, call `submit()` with a JSON string containing:
```json
{
  "type_of_statistic": "z | t | F | chi2 | r",
  "test_statistic": <numeric value>,
  "degrees_of_freedom_1": <numeric value or null>,
  "degrees_of_freedom_2": <numeric value or null>,
  "sample_size": <integer>,
  "p_value": <numeric value or null>,
  "conclusion": "same conclusion as original study | no effect or inconclusive | opposite effect",
  "reasoning": "<your conclusion in 1-3 sentences, referencing the actual numbers>",
  "dependent_variable": "<dependent variable or outcome used>",
  "main_predictor": "<main predictor / treatment / grouping variable used>",
  "sample_definition": "<which observations were included/excluded>",
  "model_specification": "<statistical model or test used>",
  "controls": "<controls/covariates used, or none>",
  "operationalization_notes": "<brief rationale for these analytic choices>"
}
```
\end{lstlisting}

\medskip
For RQ3, the \emph{Materials} block above is replaced as follows. Redacted-methods variant:

\begin{lstlisting}[style=promptstyle]
## Materials

Data directory: `<data dir>`
Original paper (methods/research-design section redacted): Markdown file `<paper path>`.

The paper's introduction, results, and discussion are visible, but the methods / data / research-design section has been removed and replaced with a short redaction marker. You must therefore decide the operationalization (subsample, dependent variable, model, controls) yourself. The paper may report results numbers for this claim - do not copy them; run the analysis on the data and report the statistic your code produces. Tables are available as readable Markdown files (`tbl-N.md`) in the same `<paper dir>/` directory. Figures referenced inline in the paper Markdown (e.g. `![img-N.jpeg](img-N.jpeg)`) are present in `<paper dir>/` - use the `view_image` tool to look at one.
\end{lstlisting}

\medskip
Abstract-only variant:

\begin{lstlisting}[style=promptstyle]
## Materials

Data directory: `<data dir>`
Original paper (abstract only): Markdown file `<paper path>`.

Only the title block and abstract are available - you do NOT have the methods, data description, results, or discussion. You must decide the operationalization (subsample, dependent variable, model, controls) yourself based on the claim and the structure of the data.
\end{lstlisting}

\medskip
\noindent\textbf{Perspective instructions}

The two perspective instructions are reported below: 
\medskip
\noindent\textbf{Confirmatory:}
\begin{lstlisting}[style=promptstyle]
Approach this analysis as a supportive analyst who expects this claim to be empirically robust and to hold up under reanalysis. Follow the most direct analysis implied by the available materials. If a necessary analytic choice remains ambiguous, resolve it in the way that provides the most charitable defensible test of the claim while still remaining fair. Commit to this one analysis and report what it shows.
\end{lstlisting}

\medskip
\noindent\textbf{Critical:}
\begin{lstlisting}[style=promptstyle]
Approach this analysis as a skeptical analyst who questions whether this claim is as robust as the authors assert. Follow the most direct analysis implied by the available materials. If a necessary analytic choice remains ambiguous, resolve it in the way that provides the most demanding defensible test of the claim while still remaining fair. Commit to this one analysis and report what it shows.
\end{lstlisting}

\end{document}